\documentclass{article}
\usepackage{iclr2024_conference,times}

\usepackage[utf8]{inputenc} %

\usepackage{graphicx}

\usepackage[T1]{fontenc}

\IfFileExists{headers/config/showoverfull.config}{
	\overfullrule=1cm
}{
}

\usepackage{marginnote}

\usepackage[backgroundcolor=none,linecolor=red,textsize=footnotesize]{todonotes}

\usepackage{etoolbox}

\newbool{includeappendix}
\setbool{includeappendix}{true} %
\IfFileExists{headers/config/noappendix.config}{
	\setbool{includeappendix}{false}
}{}

\newif\ifincludeappendixx
\ifbool{includeappendix}{
	\includeappendixxtrue
}{
	\includeappendixxfalse
}

\usepackage{xr} %
\usepackage{filecontents}

\ifbool{includeappendix}{}{
	\input{appendix-labels-loader}

	\externaldocument{appendix-labels}
}

\definecolor{my-full-blue}{HTML}{1F77B4}

\definecolor{my-full-orange}{HTML}{FF7F0E}

\definecolor{my-full-green}{HTML}{2CA02C}

\definecolor{my-full-red}{HTML}{d62728}

\definecolor{my-full-purple}{HTML}{9467bd}

\definecolor{my-full-brown}{HTML}{8c564b}

\definecolor{my-full-pink}{HTML}{e377c2}

\definecolor{my-full-gray}{HTML}{7f7f7f}

\definecolor{my-full-olive}{HTML}{bcbd22}

\definecolor{my-full-cyan}{HTML}{17becf}

\definecolor{muted-green}{RGB}{87,132,89}
\definecolor{muted-red}{RGB}{218,90,84}

\colorlet{cgreen}{muted-green}
\colorlet{cred}{muted-red}

\colorlet{my-blue}{my-full-blue!30}
\colorlet{my-orange}{my-full-orange!30}
\colorlet{my-green}{my-full-green!30}
\colorlet{my-red}{my-full-red!30}
\colorlet{my-purple}{my-full-purple!30}
\colorlet{my-brown}{my-full-brown!30}
\colorlet{my-pink}{my-full-pink!30}
\colorlet{my-gray}{my-full-gray!30}
\colorlet{my-olive}{my-full-olive!30}
\colorlet{my-cyan}{my-full-cyan!30}

\usepackage{listings}

\usepackage{textcomp}

\usepackage{xcolor}

\usepackage[scaled=0.8]{beramono}

\definecolor{ckeyword}{HTML}{7F0055}
\definecolor{ccomment}{HTML}{3F7F5F}
\definecolor{cstring}{HTML}{2A0099}

\lstdefinestyle{numbers}{
	numbers=left,
	framexleftmargin=20pt,
	numberstyle=\tiny,
	firstnumber=auto,
	numbersep=1em,
	xleftmargin=2em
}

\lstdefinestyle{layout}{
	frame=none,
	captionpos=b,
}

\lstdefinestyle{comment-style}{
	morecomment=[l]//,
	morecomment=[s]{/*}{*/},
	commentstyle={\color{ccomment}\itshape},
}

\lstdefinestyle{string-style}{
	morestring=[b]",%
	morestring=[b]',%
	stringstyle={\color{cstring}},
	showstringspaces=false,%
}

\lstdefinestyle{keyword-style}{
	keywordstyle={\ttfamily\bfseries},
	morekeywords={
		function,
		constructor,
		int,
		bool,
		return,
		returns,
		uint
	},
	morekeywords = [2]{},
	keywordstyle = [2]{\text},
	sensitive=true,
}

\lstdefinestyle{input-encoding}{
	inputencoding=utf8,
	extendedchars=true,
	literate=
	{ℝ}{$\reals$}1%
	{→}{$\rightarrow$}1%
	{α}{$\alpha$}1%
	{β}{$\beta$}1%
	{λ}{$\lambda$}1%
	{θ}{$\theta$}1%
	{ϕ}{$\phi$}1%
}

\lstdefinestyle{escaping}{
	moredelim={**[is][\color{blue}]{\%}{\%}},
	escapechar=|,
	mathescape=true
}

\lstdefinestyle{default-style}{
	basicstyle=\fontencoding{T1}\ttfamily\footnotesize,
	style=numbers,
	style=layout,
	style=comment-style,
	style=string-style,
	style=keyword-style,
	style=input-encoding,
	style=escaping,
	tabsize=2,
	upquote=true
}

\lstdefinelanguage{BASIC}{
	language=C++,
	style=default-style
}[keywords,comments,strings]%

\usepackage{amsmath,amsfonts,bm}

\def\1{\bm{1}}

\def\va{{\bm{a}}}
\def\vb{{\bm{b}}}

\def\vh{{\bm{h}}}

\def\vl{{\bm{l}}}

\def\vu{{\bm{u}}}
\def\vv{{\bm{v}}}
\def\vw{{\bm{w}}}
\def\vx{{\bm{x}}}
\def\vy{{\bm{y}}}

\def\mA{{\bm{A}}}

\def\mW{{\bm{W}}}

\DeclareMathAlphabet{\mathsfit}{\encodingdefault}{\sfdefault}{m}{sl}
\SetMathAlphabet{\mathsfit}{bold}{\encodingdefault}{\sfdefault}{bx}{n}

\newcommand{\R}{\mathbb{R}}

\usepackage[hidelinks]{hyperref}
\usepackage{url}

\usepackage{adjustbox}
\usepackage[capitalise]{cleveref}
\usepackage{scrextend}

\usepackage{enumitem}
\usepackage{threeparttable}
\usepackage{multirow, makecell}
\usepackage{booktabs}
\usepackage{xspace}
\usepackage{wrapfig}
\usepackage{dsfont}

\usepackage{subcaption}
\usepackage{caption}
\usepackage{algorithm,algcompatible,amssymb,amsmath}

\algnewcommand\RETURN{\State \textbf{return} }
\usepackage{amsthm}
\usepackage{thmtools}
\usepackage{thm-restate}
\usepackage{mathtools}
\usepackage[makeroom]{cancel}

\declaretheoremstyle[
  spaceabove=0.5em, %
  spacebelow=0.01em,%
  headfont=\normalfont\bfseries,
  bodyfont=\normalfont,
  postheadspace=0.5em,
]{mystyle}

\declaretheoremstyle[
  spaceabove=0.5em, %
  spacebelow=0.01em,%
  headfont=\normalfont\itshape,
  bodyfont=\normalfont,
  postheadspace=0.5em,
  qed=$\square$,
]{myproofstyle}

\crefname{thm}{Theorem}{Theorems}

\crefname{lem}{Lemma}{Lemmas}

\crefname{cor}{Corollary}{Corollaries}

\crefname{defi}{Definition}{Definitions}

\declaretheorem[name={Proof}, style=myproofstyle, unnumbered]{Proof}

\usepackage{tikz}
\usetikzlibrary{calc,decorations,decorations.pathmorphing}
\usetikzlibrary{positioning,fit,arrows}
\usetikzlibrary{decorations.markings}
\usetikzlibrary{shapes,shapes.geometric}
\usetikzlibrary{shadows,patterns,snakes}
\usetikzlibrary{backgrounds,decorations.pathreplacing,automata}
\usetikzlibrary{intersections}
\usetikzlibrary{angles,quotes}
\usetikzlibrary{plotmarks}
\usetikzlibrary{patterns}
\usetikzlibrary{arrows.meta}
\usetikzlibrary{shapes.misc}
\usetikzlibrary{chains}
\usepackage{siunitx}
\newcolumntype{d}[1]{S[table-format=#1]}
\usetikzlibrary{patterns.meta}

\makeatletter
\def\extractcoord#1#2#3{
	\path let \p1=(#3) in \pgfextra{
		\pgfmathsetmacro#1{\x{1}/\pgf@xx}
		\pgfmathsetmacro#2{\y{1}/\pgf@yy}
		\xdef#1{#1} \xdef#2{#2}
	};
}
\makeatother

\newcommand{\bc}[1]{\mathcal{#1}}
\newcommand{\bs}[1]{\boldsymbol{#1}}
\DeclareMathOperator*{\relu}{ReLU}
\newcommand{\RR}{\relu}

\newcommand{\I}{\mathbb{I}\xspace}

\newcommand{\cpwl}{\operatorname{CPWL}}

\usepackage{pifont}%
\newcommand{\cmark}{\ding{51}}%
\newcommand{\xmark}{\ding{55}}%

\newcommand{\deeppoly}{\textsc{DeepPoly}\xspace}
\newcommand{\boxd}{\textsc{IBP}\xspace}
\newcommand{\ibp}{\textsc{IBP}\xspace}
\renewcommand{\triangle}{$\Delta$\xspace}

\newcommand{\mn}{\textsc{MN}\xspace}
\newcommand{\dpo}{\textsc{DP-1}\xspace}
\newcommand{\dpO}{\textsc{DP-0}\xspace}

\newcommand{\acnfc}{\protecting{\tikz[]{\draw[-, style=dashed](-0.1, -0.01) -- (0.22, -0.01);\node[] at (0,0) {};}\xspace}}
\newcommand{\acfc}{\protecting{\tikz[]{\draw[black,thick](-0.1, -0.01) -- (0.22, -0.01);\node[] at (0,0) {};}\xspace}}
\newcommand{\neigh}{\protecting{\tikz[]{\node[circle, fill=blue!80, inner sep=0pt, outer sep=0pt, opacity=0.5, minimum size=2.3mm]{};}\xspace}}

\usepackage{comment}


\crefformat{section}{\S#2#1#3}

\crefrangeformat{section}{\S#3#1#4\crefrangeconjunction\S#5#2#6}

\crefmultiformat{section}{\S#2#1#3}{\crefpairconjunction\S#2#1#3}{\crefmiddleconjunction\S#2#1#3}{\creflastconjunction\S#2#1#3}

\newcommand{\crefrangeconjunction}{--}

\crefname{listing}{Lst.}{listings}
\crefname{line}{Lin.}{Lin.}
\crefname{appendix}{App.}{App.}

\newcommand{\appref}[1]{%
	\ifbool{includeappendix}{\cref{#1}}{the appendix}%
}
\newcommand{\Appref}[1]{%
	\ifbool{includeappendix}{\cref{#1}}{The appendix}%
}

\title{Expressivity of ReLU-Networks \\under Convex Relaxations}

\author{%
  Maximilian Baader\thanks{Equal contribution}, Mark Niklas Müller\footnotemark[1], Yuhao Mao, Martin Vechev\\
  Department of Computer Science\\
  ETH Zurich, Switzerland\\
  \texttt{\{mbaader, mark.mueller, yuhao.mao, martin.vechev\}@inf.ethz.ch} \\
}

\iclrfinalcopy %
\begin{document}

\maketitle

\begin{abstract}
	\vspace{-2mm}
	Convex relaxations are a key component of training and certifying provably safe neural networks. However, despite substantial progress, a wide and poorly understood accuracy gap to standard networks remains, raising the question of whether this is due to fundamental limitations of convex relaxations. Initial work investigating this question focused on the simple and widely used IBP relaxation. It revealed that some univariate, convex, continuous piecewise linear (CPWL) functions cannot be encoded by any ReLU network such that its IBP-analysis is precise.
To explore whether this limitation is shared by more advanced convex relaxations, we conduct the first in-depth study on the expressive power of ReLU networks across all commonly used convex relaxations. We show that: (i) more advanced relaxations allow a larger class of \emph{univariate} functions to be expressed as precisely analyzable ReLU networks, (ii) more precise relaxations can allow exponentially larger solution spaces of ReLU networks encoding the same functions, and (iii) even using the most precise single-neuron relaxations, it is impossible to construct precisely analyzable ReLU networks that express \emph{multivariate}, convex, monotone CPWL functions.

\end{abstract}

\begin{wraptable}[11]{r}{0.60 \textwidth}
    \centering
    \vspace{-4.7mm}
    \caption{
        Expressivity of different relaxations. 
        Novel results are {\color{cred} red \xmark} and {\color{cgreen} green \cmark}. 
        Previous results are in black (\xmark, \textbf{?}). 
        M: monotone, C: convex, MC: monotone and convex.
    }
    \vspace{-2.5mm}
    \renewcommand{\arraystretch}{1.0}
    \label{tab:overview}
    \scalebox{0.80}{
        \begin{tabular}{@{}llcccc@{}}
            \toprule
            $\bc{X}$ & Relaxation & CPWL & M-CPWL & C-CPWL & MC-CPWL \\
            \midrule
            \multirow{6}{*}{$\mathbb{R}$} & \boxd & \xmark & {\color{cgreen}\cmark} & \xmark & {\color{cgreen}\cmark} \\
            & \deeppoly-0 & \textbf{?} & {\color{cgreen}\cmark} & {\color{cgreen}\cmark} & {\color{cgreen}\cmark} \\
            & \deeppoly-1 & \textbf{?} & {\color{cgreen}\cmark} & {\color{cgreen}\cmark} & {\color{cgreen}\cmark} \\
            & \triangle & \textbf{?} & {\color{cgreen}\cmark} & {\color{cgreen}\cmark} & {\color{cgreen}\cmark} \\
            & Multi-Neuron$_\infty$ & {\color{cgreen}\cmark} & {\color{cgreen}\cmark} & {\color{cgreen}\cmark} & {\color{cgreen}\cmark} \\
            \midrule
            $\mathbb{R}^d$ & \triangle & {\color{cred}\xmark} & {\color{cred}\xmark} & {\color{cred}\xmark} & {\color{cred}\xmark} \\
            \bottomrule
        \end{tabular}
    }
    \vspace{-3mm}
\end{wraptable}

\vspace{-3mm}
\section{Introduction} \label{sec:introduction}
\vspace{-1mm}
With the increased deployment of neural networks in mission-critical applications, formal robustness guarantees against adversarial examples \citep{BiggioCMNSLGR13,SzegedyZSBEGF13} have become an important and active field of research.
Many popular certification methods \citep{ZhangWCHD18,SinghGMPV18,SinghGPV19B,SinghGPV19} provide such safety guarantees by using convex relaxations to compute over-approximations of a network's reachable set w.r.t. an adversary specification. 
However, despite significant progress, a wide and poorly understood accuracy gap between robust and conventional networks remains. This raises the fundamental question: 
\begin{center}
    \vspace{-.5mm}
    \vspace{.5mm}
    \emph{Is the expressivity of ReLU-networks under convex relaxations fundamentally limited?}
    \vspace{-1.0mm}
\end{center}
Investigating this question, \citet{MirmanBV22} prove that, the class of convex, continuous-piecewise-linear (CPWL) functions \emph{cannot} be encoded as ReLU-networks such that their analysis with the simple \ibp-relaxation \citep{GehrMDTCV18, GowalIBP2018}, is \emph{precise}.

\vspace{-1.5mm}
\paragraph{This Work: Expressivity of Common Relaxations}
To investigate whether this limitation of \ibp is fundamental to all single-neuron convex relaxations, we conduct the first in-depth study on the expressive power of ReLU networks under \emph{all} commonly used relaxations. 
To this end, we consider CPWL functions, naturally represented by ReLU networks, and two common restrictions, convexity and monotonicity.
We illustrate our key findings in \cref{tab:overview}, showing novel results as {\color{cred} red \xmark} and {\color{cgreen} green \cmark} and previous results (\xmark, \cmark) and open questions (\textbf{?}) in black. 

\textbf{Key Results on Univariate Functions} \hspace{1em}In this work, we prove the following key results:
\vspace{-1.5mm}
\begin{itemize}
    \renewcommand{\itemsep}{0.2mm}
    \item The most precise single-neuron relaxation, \triangle (\citet{WongK18}), and the popular \deeppoly-relaxation \citep{SinghGPV19,ZhangWCHD18} do not share \ibp's limitation and can express univariate, \emph{convex}, CPWL functions precisely.
    \item All considered relaxations, including \ibp, can express univariate, \emph{monotone}, CPWL functions precisely. 
    \item  The \triangle-relaxation permits an exponentially larger network solution space for convex CPWL functions compared to the less precise \deeppoly-relaxation.
    \item Multi-neuron relaxations \citep{SinghGPV19B,MullerMSPV22} can express all univariate, CPWL functions precisely using a single layer.
\end{itemize}

Having thus shown that, for \emph{univariate functions}, the expressivity of ReLU networks under convex relaxations \emph{is not fundamentally limited}, we turn our analysis to multivariate functions. 

\textbf{Key Results on Multivariate Functions} \hspace{1em} In this setting, we prove the following result:
\vspace{-1.5mm}
\begin{itemize}
    \item No single-neuron convex relaxation can precisely express even the heavily restricted class of multivariate, convex, monotone, CPWL functions.
\end{itemize}
Interestingly, the exact analysis of such monotone functions on box input regions is trivial, making the failure of convex relaxations even more surprising.
In fact, CPWL functions as simple as $f(x,y) = \max(x,y) = y + \RR(x-y)$ cannot be encoded by any finite ReLU network such that its \triangle-analysis is precise. 
We thus conclude that, for \emph{multivariate functions}, the expressivity of ReLU networks under single-neuron convex relaxations \emph{is fundamentally limited}.

\textbf{Implications of our Results for Certified Training}
While we believe our results to be of general interest, they have particularly interesting implications for certified training. 
In this area, all state-of-the-art methods \citep{MullerE0V23,MaoMFV23,PalmaBDKSL23} are based on the simple \ibp-relaxation even though it induces strong regularisation which severely reduces accuracy.
While \citet{JovanovicBBV22} show that more precise relaxations induce significantly harder optimization problems, it remains an open question whether solving these would actually yield networks with better performance. Our results represent a major step towards answering this question.

Specifically in the univariate setting, we show that more precise relaxations increase expressivity (see \cref{tab:overview}) and lead to larger network solution spaces (compare \cref{thm:deeppoly_convex,lem:triangle_conv}). Thus, we hypothesize that using them during training yields a larger effective hypothesis space for the same network architecture. Importantly, this implies that networks with higher performance could indeed be obtained if we can overcome the optimization issues described by \citet{JovanovicBBV22}.

However, in the multivariate setting, perhaps surprisingly, we show that even the most precise single-neuron relaxations severely limit expressivity (see \cref{cor:triangle_impossibility}). This highlights the need for further study of more precise analysis methods such as multi-neuron or non-convex relaxations.

\section{Background on Convex Relaxations}

Below, we first discuss notation and background before defining key concepts.

\paragraph{Notation}
We denote vectors with bold lower-case letters $\va \in \R^n$, matrices with bold upper-case letters $\mA \in \R^{n \times d}$, and sets with upper-case calligraphic letters $\bc{A} \subset \R$.
We refer to a hyperrectangle $\bc{B} \subset \R^n$ as a box. Further, we consider (finite) ReLU networks $h$ with arbitrary skip connections.

\subsection{Convex Relaxations in Neural Network Certification}\label{sec:background_convex}

Here, we discuss neural network certification methods based on convex relaxations. These methods cast the robustness problem as an optimization problem and make it tractable by replacing the non-convex activation functions with convex relaxations in their input-output space. Different choices of relaxations lead to different trade-offs between precision and computational cost.
Generally, all considered convex relaxations allow us to compute linear bounds on the output of a real-valued neural network $h$ 
\begin{equation*}
    \{\mA_{l_i} \vx + b_{l_i}\}_{i \in \bc{L}} \leq h(\vx) \leq \{ \mA_{u_j} \vx + b_{u_j} \}_{j \in \bc{U}},
\end{equation*}
given an input region $\bc{B} = [\vl, \vu] \ni \vx$, where $\bc{L}$ and $\bc{U}$ are some index sets. These bounds can in-turn be bounded by $\vl_y = \min_{\vx \in \bc{B}} \max_{i \in \bc{L}}(\mA_{l_i} \vx + b_{l_i}) \in \R$ and $\vu_y = \max_{\vx \in \bc{B}} \min_{j \in \bc{U}}(\mA_{u_j} \vx + b_{u_j}) \in \R$. Hence, we have $\vl_y \leq \vh(\vx) \leq \vu_y$.

\paragraph{IBP}
\begin{wrapfigure}[10]{r}{0.40 \textwidth}
	\centering
	\vspace{-6mm}
	\scalebox{0.90}{\begin{tikzpicture}
	\draw[->] (-2.5, 0) -- (2.5, 0) node[right,scale=0.85] {$v$};
	\draw[->] (0, -0.4) -- (0, 2.0) node[above,scale=0.85] {$y$};

	\def\a{-1.8}
	\def\b{1.4}
	\coordinate (a) at ({\a},{0});
	\coordinate (b) at ({\b},{\b});
	\coordinate (c) at ({0},{0});
	\coordinate (d) at ({\a},{\b});
	\coordinate (e) at ({\b},{0});
	
	\node[circle, fill=black, minimum size=3pt,inner sep=0pt, outer sep=0pt] at (a) {};
	\node[circle, fill=black, minimum size=3pt,inner sep=0pt, outer sep=0pt] at (b) {};
	\node[circle, fill=black, minimum size=3pt,inner sep=0pt, outer sep=0pt] at (c) {};
	
	\fill[fill=blue!90,opacity=0.3] (a) -- (e) -- (b) -- (d) -- cycle;
	\draw[black,thick] (a) -- (c) -- (b);
	\draw[-] (b) -- ($(b)+(0.3,0.3)$);
	\draw[-] ({\a},-0.4) -- ({\a},0.4);
	\draw[-] ({\b},-0.4) -- ({\b},1.9);
	
	\node[anchor=south west,align=center,scale=0.85] at ({\a},-0.50) {$l$};
	\node[anchor=south west,align=center,scale=0.85] at ({\b},-0.50) {$u$};

\end{tikzpicture}}
	\vspace{-2mm}
	\caption{\ibp-relaxation (blue) of a ReLU with bounded inputs $v\in [l,u]$.}
	\label{fig:ReLU_box}
\end{wrapfigure}
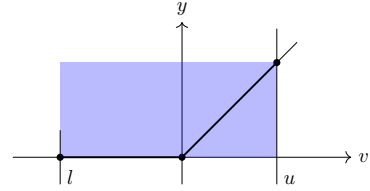
Interval bound propagation \citep{MirmanGV18, GehrMDTCV18, GowalIBP2018} only considers elementwise, constant bounds of the form $\vl \leq \vv \leq \vu$. Affine layers $\vy = \mW \vv +\vb$ are thus also relaxed as
\begin{equation*}
    \resizebox{.99\linewidth}{!}{$\tfrac{\mW (\vl+\vu) - |\mW|(\vu-\vl)}{2} + \vb \leq \mW \vv +\vb \leq \tfrac{\mW (\vl+\vu) + |\mW|(\vu-\vl)}{2} +\vb,$}
\end{equation*}
where $|\cdot|$ the elementwise absolute value. ReLU functions are relaxed by their concrete lower and upper bounds $\RR(\vl) \leq \RR(\vv) \leq \RR(\vu)$, illustrated in \cref{fig:ReLU_box}.

\paragraph{DeepPoly (DP)}
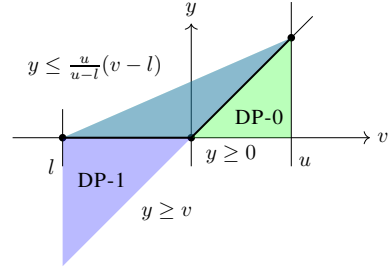
\begin{wrapfigure}[15]{r}{0.40 \textwidth}
	\centering
	\vspace{-6mm}
	\scalebox{0.95}{\begin{tikzpicture}
	\draw[->] (-2.5, 0) -- (2.5, 0) node[right,scale=0.85] {$v$};
	\draw[->] (0, -0.4) -- (0, 1.6) node[above,scale=0.85] {$y$};

	\def\a{-1.8}
	\def\b{1.4}
	\def\al{0.15}
	\coordinate (a) at ({\a},{0});
	\coordinate (a1) at ({\a},{\a});
	\coordinate (b0) at ({\b},{0});
	\coordinate (b) at ({\b},{\b});
	\coordinate (c) at ({0},{0});
	\coordinate (aub) at ({\a},{\b-(\b-\a)*\al});
	\coordinate (bub) at ({\b},{(\b-\a)*\al});
	\coordinate (alb) at ({\a},{\a*\al});
	\coordinate (blb) at ({\b},{\b*\al});
	
	\node[circle, fill=black, minimum size=3pt,inner sep=0pt, outer sep=0pt] at (a) {};
	\node[circle, fill=black, minimum size=3pt,inner sep=0pt, outer sep=0pt] at (b) {};
	\node[circle, fill=black, minimum size=3pt,inner sep=0pt, outer sep=0pt] at (c) {};
	
	\fill[fill=green!90, opacity=0.3] (a) -- (b) -- (b0) -- cycle;
	\fill[fill=blue!90, opacity=0.3] (a1) -- (a) -- (b) -- cycle;
	\draw[black,thick] (a) -- (c) -- (b);
	\draw[-] (b) -- ($(b)+(0.3,0.3)$);
	\draw[-] ({\a},-0.4) -- ({\a},0.4);
	\draw[-] ({\b},-0.4) -- ({\b},1.9);

	\node[anchor=south east,align=center,scale=0.85] at (-0.8,-0.8) {\dpo};
	\node[anchor=north west,align=center,scale=0.85] at (0.5,0.5) {\dpO};

	\node[anchor=north east,align=center,scale=0.85] at ({\a},-0.20) {$l$};
	\node[anchor=south west,align=center,scale=0.85] at ({\b},-0.50) {$u$};	
	\node[anchor=south east,align=center,scale=0.85] at ({-0.3},0.70) {$y\leq \frac{u}{u-l} (v-l)$};
	\node[anchor=north west,align=center,scale=0.85] at ({0.1},0.00) {$y\geq 0$};
	\node[anchor=north west,align=center,scale=0.85] at (-0.8,-0.8) {$y\geq v$};

\end{tikzpicture}}
	\vspace{-2mm}
	\caption{\deeppoly-1 (blue) and \deeppoly-0 (green) abstraction of a ReLU with bounded inputs $v \in [l,u]$.}
	\label{fig:ReLU_deeppoly}
\end{wrapfigure}

DeepPoly, introduced by \citet{SinghGPV19}, is mathematically identical to CROWN \citep{ZhangWCHD18} and based on recursively deriving linear bounds of the form
\begin{equation*}
    \mA_l \vx + \va_l \leq \vv \leq \mA_u \vx + \va_u
\end{equation*}
on the outputs of every layer. While this allows affine layers to be handled exactly, ReLU layers $\vy = \RR(\vv)$ are relaxed neuron-wise, using one of the two relaxations illustrated in \cref{fig:ReLU_deeppoly}
\begin{equation*}
    \boldsymbol{\lambda} \vv \leq \RR(\vv) \leq (\vv-\vl) \frac{\vu}{\vu - \vl},
\end{equation*}
where product and division are elementwise.
Typically, the lower-bound slope $\lambda \in \{0,1\}$ is chosen depending on the input bounds $l$ and $u$. In this work, however, we analyze the relaxations obtained by always choosing the same lower-bound, which we denote with \deeppoly-0 (\dpO, green in \cref{fig:ReLU_deeppoly}) and \deeppoly-1 (\dpo, blue). 

\paragraph{Triangle-Relaxation (\triangle)}
\begin{wrapfigure}[8]{r}{0.40 \textwidth}
	\centering
	\vspace{-5mm}
	\scalebox{0.95}{\begin{tikzpicture}
	\draw[->] (-2.5, 0) -- (2.5, 0) node[right,scale=0.85] {$v$};
	\draw[->] (0, -0.4) -- (0, 1.6) node[above,scale=0.85] {$y$};

	\def\a{-1.8}
	\def\b{1.4}
	\def\al{0.15}
	\coordinate (a) at ({\a},{0});
	\coordinate (a1) at ({\a},{\a});
	\coordinate (b0) at ({\b},{0});
	\coordinate (b) at ({\b},{\b});
	\coordinate (c) at ({0},{0});
	\coordinate (aub) at ({\a},{\b-(\b-\a)*\al});
	\coordinate (bub) at ({\b},{(\b-\a)*\al});
	\coordinate (alb) at ({\a},{\a*\al});
	\coordinate (blb) at ({\b},{\b*\al});
	
	\node[circle, fill=black, minimum size=3pt,inner sep=0pt, outer sep=0pt] at (a) {};
	\node[circle, fill=black, minimum size=3pt,inner sep=0pt, outer sep=0pt] at (b) {};
	\node[circle, fill=black, minimum size=3pt,inner sep=0pt, outer sep=0pt] at (c) {};
	
	\fill[fill=blue!90, opacity=0.3] (a) -- (c) -- (b) -- cycle;
	\draw[black,thick] (a) -- (c) -- (b);
	\draw[-] (b) -- ($(b)+(0.3,0.3)$);
	\draw[-] ({\a},-0.4) -- ({\a},0.4);
	\draw[-] ({\b},-0.4) -- ({\b},1.9);

	\node[anchor=north east,align=center,scale=0.85] at ({\a},-0.20) {$l$};
	\node[anchor=south west,align=center,scale=0.85] at ({\b},-0.50) {$u$};	
	\node[anchor=south east,align=center,scale=0.85] at ({-0.3},0.70) {$y\leq \frac{u}{u-l} (v-l)$};
	\node[anchor=north west,align=center,scale=0.85] at (-{1.2},0.00) {$y\geq 0$};
	\node[anchor=north west,align=center,scale=0.85] at (0.4,0.6) {$y\geq v$};

\end{tikzpicture}}
	\vspace{-0mm}
	\caption{\triangle-relaxation (blue) of a ReLU with bounded inputs $v\in [{l},{u}]$.}
	\label{fig:ReLU_triangle}
\end{wrapfigure}
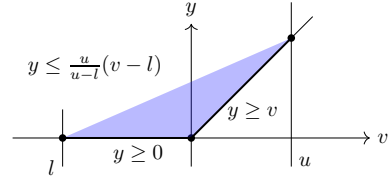

In contrast to the above convex relaxations, the \triangle-relaxation \citep{WongK18,DvijothamSGMK18,SalmanY0HZ19,QinDOBSGUSK19} maintains multiple linear upper- and lower-bounds on every network activation $v$. We write
\begin{equation*}
    \begin{rcases} \mA_{l_1} \vx + \va_{l_1}, \\ \qquad\;  \vdots \\ \mA_{l_n} \vx + \va_{l_n}, \end{rcases} \leq \vv \leq \begin{cases} \mA_{u_1} \vx + \va_{u_1}, \\ \qquad\;  \vdots \\ \mA_{u_n} \vx + \va_{u_n}. \end{cases}
\end{equation*}
Unstable ReLU activation $\vy=\RR(\vv)$ with $\vl < \mathbf{0} < \vu$ are relaxed with their convex hull as illustrated in \cref{fig:ReLU_triangle}
\begin{equation*}
    \begin{rcases} \mathbf{0} \\ \vv \end{rcases} \leq \RR(\vv) \leq (\vv-\vl) \frac{\vu}{\vu - \vl},
\end{equation*} 
where again, product and division are elementwise.
Note that this can lead to an exponential growth (with the depth of the network) in the number of constraints for any given activation.

\paragraph{Multi-Neuron Relaxation (\mn)}
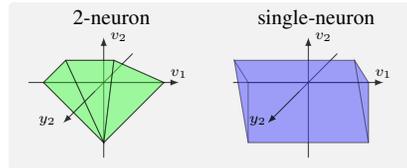
\begin{wrapfigure}[11]{r}{0.40\textwidth}
	\centering
	\vspace{-4mm}
	\scalebox{0.8}{\begin{tikzpicture}
	\tikzset{>=latex}
	
	\def\dx{1.7}
	\def\yo{-1.5}
	
	\colorlet{csingle}{blue!70}
	\colorlet{cmulti}{green!70}

	\node
	[fill=black!05, rectangle, rounded corners=2pt,
	minimum width=6.8cm, minimum height=2.8cm,
	anchor=north
	] at (0, 0) {
	};
	\node [anchor=north,align=center,] at (-\dx,0.0) { 2-neuron};
	\node [anchor=north,align=center,] at (\dx,0.0) { single-neuron};

	\node (c2l) at (-\dx, \yo) {
		\begin{tikzpicture}[scale=0.5]
			
			\coordinate (O) at (0, 0, 0);
			\coordinate (tc) at (2, 0, 0);
			\coordinate (bc) at (-2, 0, 0);
			\coordinate (mn) at (0, -2, 0);
			\coordinate (me) at (0, 0, 2);
			\coordinate (mse) at (0.8, 1.2, 1.2);
			\coordinate (mw) at (0, 0, -2);
			\coordinate (bse) at (-0.8, 1.2,1.2);

			\coordinate (xp) at (2.5, 0, 0);
			\coordinate (xn) at (-2.5, 0, 0);
			\coordinate (yp) at (0, 1.5, 0);
			\coordinate (yn) at (0, -2.5, 0);
			\coordinate (zp) at (0, 0, 3.5);
			\coordinate (zn) at (0, 0, -2.5);
			\coordinate (zm) at (0, 0, {2*1.2/3.2});

			\draw[-,opacity=0.8] (zn) -- (zm) ;
			\draw[->,opacity=0.8] (xn) -- (xp) ;
			\draw[->,opacity=0.8] (yn) -- (yp) ;
			
			\draw[fill=cmulti, opacity=0.3] (bc) -- (mn) -- (tc) -- cycle;
			\draw[fill=cmulti, opacity=0.3] (bc) -- (bse) -- (mse) -- (tc) -- cycle;
			\draw[fill=cmulti, opacity=0.3] (bc) -- (bse) -- (mn) -- cycle;
			\draw[fill=cmulti, opacity=0.3] (mse) -- (bse) -- (mn) -- cycle;
			\draw[fill=cmulti, opacity=0.3] (tc) -- (mse) -- (mn) -- cycle;
			
			\draw[opacity=0.5] (tc) -- (mn);  
			\draw[opacity=0.5] (tc) -- (mse);
			\draw[opacity=0.5] (bc) -- (mn);  
			\draw[opacity=0.5] (bc) -- (bse);
			\draw[opacity=0.5] (bse) -- (mse);  
			\draw[opacity=0.5] (mn) -- (mse);
			\draw[opacity=0.5] (mn) -- (bse);

		\draw[->,opacity=0.8] (zm) -- (zp) ;
			
		\node [font=\scriptsize, align=center, scale=1.] at (2.5,0.3,0) {$v_1$};
		\node [font=\scriptsize, align=center, scale=1.] at (0.5,1.5,0) {$v_2$};
		\node [font=\scriptsize, align=center, scale=1.] at (-0.3,0.3,4.1) {$y_2$};
			
		\end{tikzpicture}
	};

	\node (c2l) at (\dx, \yo) {
	\begin{tikzpicture}[scale=0.5]
		
		\coordinate (O) at (2, 0, 0);
		\coordinate (bc) at (-2, 0, 0);
		\coordinate (mn) at (-2, -2, 0);
		\coordinate (mse) at (2, 1.2, 1.2);
		\coordinate (mw) at (2, -2, 0);
		\coordinate (bse) at (-2, 1.2,1.2);
		
		\coordinate (xp) at (2.5, 0, 0);
		\coordinate (xn) at (-2.5, 0, 0);
		\coordinate (yp) at (0, 1.5, 0);
		\coordinate (yn) at (0, -2.5, 0);
		\coordinate (zp) at (0, 0, 3.5);
		\coordinate (zm) at (0, 0, {2*1.2/3.2});
		\coordinate (zn) at (0, 0, -2.5);
		
		\draw[->,opacity=0.8] (xn) -- (xp) ;
		\draw[->,opacity=0.8] (yn) -- (yp) ;
		\draw[-,opacity=0.8] (zn) -- (zm) ;

		\draw[fill=csingle, opacity=0.3] (bc) -- (bse) -- (mn) -- cycle;
		\draw[fill=csingle, opacity=0.3] (bc) -- (mn) -- (mw) -- (O) -- cycle;
		\draw[fill=csingle, opacity=0.3] (bc) -- (bse) -- (mse) -- (O) -- cycle;
		\draw[fill=csingle, opacity=0.3] (mse) -- (bse) -- (mn) -- (mw) -- cycle;
		\draw[fill=csingle, opacity=0.3] (O) -- (mse) -- (mw) -- cycle;

		\draw[->,opacity=0.8] (zm) -- (zp) ;
		
		\node [font=\scriptsize, align=center, scale=1.] at (2.5,0.3,0) {$v_1$};
\node [font=\scriptsize, align=center, scale=1.] at (0.5,1.5,0) {$v_2$};
\node [font=\scriptsize, align=center, scale=1.] at (-0.1,0.3,4.1) {$y_2$};

	\end{tikzpicture}
};
	
\end{tikzpicture}}
	\vspace{-4mm}
	\caption{Comparison of a 2-neuron (green) and single-neuron (blue) relaxation projected into $y_2$-$v_1$-$v_2$-space for ReLU activations $y_i = \RR(v_i)$.}
	\label{fig:precision_comp}
\end{wrapfigure}
All methods introduced so far relax activation functions neuron-wise and are thus limited in precision by the (single neuron) convex relaxation barrier \citep{SalmanY0HZ19}, i.e., the activation function's convex hull in their input-output space. 

Multi-neuron relaxations, in contrast, compute the convex hull in the joint input-output space of multiple neurons in the same layer \citep{SinghGPV19B,MullerMSPV22}, or consider multiple inputs jointly \citep{TjandraatmadjaA20}. We illustrate the increase in tightness in \cref{fig:precision_comp} for a group of just $k=2$ neurons.

\subsection{Definitions}

We now define the most important concepts for this work. 

\begin{restatable}[CPWL]{defi}{cpwl_function}\label{def:cpwl_function}
    We denote the set of continuous piecewise linear functions $f \colon \bc{X} \to \bc{Y}$ by $\cpwl(\bc{X}, \bc{Y})$. Further, if $\bc{X}$ is some interval $\I \subset \R$, then we enumerate the points where $f$ changes slope and call them $x_i$, where $0 \leq i \leq n$, $i < j$ implies $x_i < x_j$, and $\bc{X} = [x_0, x_n]$. 
\end{restatable}

All $\cpwl$ functions $f\colon \I \to \R$ satisfy $f(x) = f(x_i) + (x - x_i) \tfrac{f(x_{i+1}) - f(x_i)}{x_{i+1} - x_{i}}$ for $x \in [x_i, x_{i+1}]$. We denote by M-CPWL, C-CPWL, and MC-CPWL the class of monotone (M), convex (C), and monotone \& convex (MC) CPWL functions, respectively. 
Next, we define the encoding of a function $f$ by a network $h$:

\begin{restatable}[Encoding]{defi}{encoding}\label{thm:encoding}
    A neural network $h\colon \bc{X} \to \bc{Y}$ \emph{encodes} a function $f\colon \bc{X} \to \bc{Y}$ if and only if for all $x \in \bc{X}$ we have $h(x) = f(x)$.
\end{restatable}

In the following, $D$ denotes a convex relaxation and can be \boxd, \deeppoly-0 (\dpO), \deeppoly-1 (\dpo), \triangle, or Multi-Neuron (\mn):

\begin{restatable}[Analysis]{defi}{analysis} \label{def:analysis}
    Let $h\colon \bc{X} \!\to\! \bc{Y}$ be a network, $D$ a convex relaxation, and $\bc{B} \subset \bc{X}$ an input box. We denote by $h^D(\bc{B})$ the polytope in $h$'s input-output space containing the graph $\{(\vx, h(\vx)) \mid \vx \in \bc{B}\} \!\subseteq\! h^D(\bc{B}) \!\subseteq\! \bc{X} \times \bc{Y}$ of $h$ on $\bc{B}$, as obtained with $D$ and refer to it as the $D$-\emph{analysis} of $h$ on $\bc{B}$. 
\end{restatable}

For $\bc{Y} \subseteq \R$, we denote the interval bounds of $f$ on $\bc{B}$ by $[\underline{f(\bc{B})}, \overline{f(\bc{B})}] := [\min_{\vx \in \bc{B}} f(\vx), \max_{\vx \in \bc{B}} f(\vx)]$ and, similarly, the interval bounds implied by $h^D(\bc{B})$ as $[\underline{h^D(\bc{B})}, \overline{h^D(\bc{B})}] := [\min_{(\vx,y) \in h^D(\bc{B})} y, \max_{(\vx,y) \in h^D(\bc{B})} y]$. 

As any $D$-analysis of $h$ captures the set of all possible outputs $h(\vx), \vx \in \bc{B}$, it is of key interest to us to investigate when the analysis does not lose precision. 
Specifically, whether the linear output bounds $h^D(\bc{B})$ do not exceed the interval bounds of $f$ on $\bc{B}$ anywhere on $\bc{B}$:

\begin{restatable}[Precise]{defi}{precise_analysis}\label{def:precise_analysis}
    Let $h$ be a network encoding $f \colon \bc{X} \to \bc{Y}$ and $D$ a convex relaxation. We say that the $D$-analysis is \emph{precise} for $h$ if it yields precise lower and upper bounds, that is for all boxes $\bc{B} \subset \bc{X}$ we have that $[\underline{h^D(\bc{B})}, \overline{h^D(\bc{B})}] = [\underline{f(\bc{B})}, \overline{f(\bc{B})}]$. 
\end{restatable}

In this work, we investigate the expressivity of ReLU-networks, that is, which function class they can encode such that their $D$-analysis is precise. Specifically:

\begin{restatable}[Expressivity]{defi}{expressiveness} \label{def:expressiveness}
    Let $D$ be a convex relaxation, $\bc{F}$ a set of functions, and $\bc{N}$ a set of networks. We say that $\bc{N}$ can $D$-\emph{express} $\bc{F}$ precisely, if and only if, for all $f \in \bc{F}$, there exists a network $h \in \mathcal{N}$, such that $h$ encodes $f$ and its $D$-analysis is precise. 
\end{restatable}   

We can often replace (sub-)networks to encode the same function but yield a (strictly) more precise analysis in terms of the obtained input-output polytope:

\begin{restatable}[Replacement]{defi}{replacement} \label{def:replacement}
    Let $h$ and $h'$ be $\RR$ networks, $\bc{B}$ a box, and $D$ some convex relaxation. We say $h'$ can \emph{replace} $h$ with respect to $D$, if $h'^D(\bc{B}) \subseteq h^D(\bc{B})$ for all $\bc{B}$ and write $h \overset{D}{\rightsquigarrow} h'$.
\end{restatable}

\section{Related Work}

Below, we give a brief overview of the most relevant related work. 

\paragraph{Expressing CPWL Functions}
\citet{HeLXZ20} show that ReLU networks require at least 2 layers to encode CPWL functions in $\R^d$ (for $d \geq 2$) with $\lceil \log_2(d+1) \rceil$ layers always being sufficient.

\paragraph{Expressivity with \boxd}
\citet{BaaderMV20} show that for any continuous function $f \colon \Gamma \subset \R^n \to \R$ over a compact domain $\Gamma$ and $\epsilon > 0$, there exists a finite ReLU network $h$, such that its \boxd-analysis for any input box $\bc{B} \subset \Gamma$, denoted by $h^\boxd(\bc{B})$, is \emph{precise up to an $\epsilon$-error}:
\begin{equation*}
    [\underline{f(\bc{B})}+ \epsilon, \overline{f(\bc{B})}- \epsilon]
    \subseteq  
    h^\boxd(\bc{B})
    \subseteq
    [\underline{f(\bc{B})}- \epsilon, \overline{f(\bc{B})}+ \epsilon].
\end{equation*} 
An equivalent result immediately follows for all strictly more precise domains such as \dpO, \triangle, and \mn. 
\citet{WangAPJ22} propose a more efficient construction, generalize this result to squashable activation functions, and provide first results on the hardness of constructing such networks. 

Investigating what class of functions allows for an \emph{exact} \boxd-analysis, \citet{MirmanBV22} show that for any function with points of non-invertibility, i.e., $x = 0$ for $f(x) = |x|$, there does not exist a ReLU network \boxd-expressing this function. 

\paragraph{Certified Training}
Certified training methods typically compute and then optimize an upper bound on the worst-case loss over some adversary specification computed via convex relaxations. Surprisingly, using the imprecise \ibp-relaxation \citep{MirmanGV18,GowalIBP2018} consistently yields better performance than tighter relaxations \citep{WongSMK18,ZhangCXGSLBH20,BalunovicV20}. 
\citet{JovanovicBBV22} investigate this paradox and identify two key properties of the worst-case loss approximation, continuity and sensitivity, required for effective optimization, with only \boxd possessing both.
However, the heavy regularization that makes \ibp trained networks amenable to certification also severely reduces their standard accuracy.

\paragraph{Neural Network Certification} 
We distinguish complete certification methods, which, given sufficient time, can decide any property, i.e., always compute precise bounds, and incomplete methods, which sacrifice precision for speed.
\citet{SalmanY0HZ19} unify a range of incomplete certification methods including \ibp, \deeppoly, and \triangle, and show that their precision is limited by that of the \triangle-relaxation. They observe that for a wide range of networks and even when using the \triangle-relaxation, a substantial certification gap between the upper- and lower-bounds on robust accuracy remains.
Semidefinite programming based methods \citep{DathathriDKRUBS20,RaghunathanSL18} increase tightness at the cost of computational efficiency.

Early, complete certification methods directly leveraged off-the-shelf SMT \citep{KatzBDJK17,Ehlers17} or MILP solvers \citep{DuttaJST18,TjengXT19}, limiting their applicability to small networks. To improve scalability, \citet{BunelLTTKK20} formulate a branch-and-bound (BaB) framework, that recursively splits the certification problem into easier subproblems until they can be decided by cheap incomplete methods. This concept has been widely adopted and improved using more efficient solvers \citep{XuZ0WJLH21,WangZXLJHK21} and tighter constraints \citep{PalmaBBTK21,FerrariMJV22,ZhangWXLLJHK22}.

\section{Convex Relaxations for Univariate Functions}

In this section, we differentiate all convex relaxations that are commonly used for neural network certification (\ibp, \dpO, \dpo, \triangle, and \mn) in terms of their expressivity, i.e., with respect to the function classes they can analyze precisely when encoded by a ReLU network.

We first show that finite-depth ReLU networks can \ibp-express M-CPWL functions precisely (\cref{lem:box_precise_monotone}). 
This construction can be applied directly to the strictly more precise \dpO and \triangle relaxation and with slight modification also to \dpo. 
We, then, show that while finite ReLU networks can both \dpO- and \triangle-express M-CPWL and C-CPWL functions, the solution space is exponentially larger when using the more precise \triangle-analysis. 
Finally, we show that single-layer ReLU networks can \mn-express arbitrary CPWL functions. 
We defer all proofs and supplementary lemmata to \cref{app:theory_single}.

\subsection{Box}

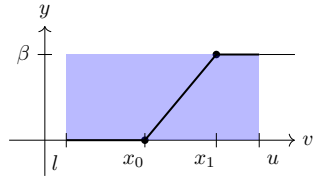
\begin{wrapfigure}[8]{r}{0.40\textwidth}
	\centering
	\vspace{-5mm}
	\scalebox{0.95}{\begin{tikzpicture}
	\draw[->] (-0.5, 0) -- (3.5, 0) node[right,scale=0.85] {$v$};
	\draw[->] (0, -0.5) -- (0, 1.6) node[above,scale=0.85] {$y$};

	\def\xO{1.4}
    \def\xo{2.4}
	\def\b{1.2}

	\def\l{+.3}
    \def\u{3.0}

	\coordinate (a) at ({\l},{0});
    \coordinate (at) at ({\l},{\b});
	\coordinate (b) at ({\xO},{0});
	\coordinate (c) at ({\xo},{\b});
	\coordinate (d) at ({\u},{\b});
    \coordinate (db) at ({\u},{0});
	
	\node[circle, fill=black, minimum size=3pt,inner sep=0pt, outer sep=0pt] at (b) {};
	\node[circle, fill=black, minimum size=3pt,inner sep=0pt, outer sep=0pt] at (c) {};
	
    \fill[color=blue!90, opacity=0.3] (a) -- (at) -- (d) -- (db) -- cycle;
	\draw[black,thick] (a) -- (b) -- (c) -- (d);
    \draw[-] (d) -- ({3.5},{\b});

    \draw[-] ({\l},-0.1) -- ({\l},0.1);
    \node[anchor=north east,align=center,scale=0.85] at ({\l},-0.10) {$l$};

	\draw[-] (-0.1,{\b}) -- (0.1,{\b});
	\node[anchor=north west,align=center,scale=0.85] at ({\u},-0.10) {$u$};	

    \draw[-] ({\u},-0.1) -- ({\u},{0.1});
	\node[anchor=east,align=center,scale=0.85] at (-0.1, {\b}) {$\beta$};	
   
    \draw[-] ({\xo},-0.1) -- ({\xo},0.1);
    \node[anchor=north east,align=center,scale=0.85] at ({\xo+0.1},-0.10) {$x_1$};

    \draw[-] ({\xO},-0.1) -- ({\xO},0.1);
    \node[anchor=north east,align=center,scale=0.85] at ({\xO+0.1},-0.10) {$x_0$};

\end{tikzpicture}}
	\vspace{-1mm}
	\caption{\ibp-analysis of the step function $\beta - \RR( \beta-\tfrac{\beta}{x_1 - x_0}\RR(x-x_0))$.}
	\label{fig:box_step}
\end{wrapfigure}

To show that M-CPWL functions can be \boxd-expressed, we begin by constructing a step function, illustrated in (\cref{fig:box_step}), as a two-layer ReLU network that can be \boxd-expressed:

\begin{restatable}[Step Function]{lem}{StepFunction} \label{lem:step_function}
    Let $\beta \in \R_{\geq 0}$ and $f \in \cpwl(\I, \R)$ s.t. $f(x) = 0$ for $x < x_0$, $f(x) = \beta$ for $x > x_1$ and linear in between. Then, $\phi_{x_0, x_1, \beta}(x) = \beta - \RR( \beta-\tfrac{\beta}{x_1 - x_0}\RR(x-x_0))$ encodes $f$.
\end{restatable}

\pagebreak

\begin{restatable}[Precise Step]{lem}{BoxPreciseStepFunction} \label{lem:box_precise_step_function}
    The \boxd-analysis of $\phi_{x_0, x_1, \beta}$ is precise. 
\end{restatable}

Intuitively, the key to this construction is to leverage that while the \boxd-relaxation of the ReLU function does not capture any relational information, it recovers the exact output interval. By using two sequential ReLUs, we allow the inner one to cut away the output-half-space $f(x) < 0$ and the outer one to cut away the half-space $f(x) > \beta$, thus obtaining a precise analysis.

We can now construct arbitrary M-CPWL functions from these step functions, allowing us to show that they too can be \boxd-expressed:

\begin{restatable}[Precise Monotone]{thm}{BoxPreciseMonotone} \label{lem:box_precise_monotone}
    Finite $\RR$ networks can \boxd-express the set of monotone $\cpwl(\I, \R)$ functions precisely. 
\end{restatable}

\subsection{DeepPoly-0}

\begin{wrapfigure}[25]{r}{0.30 \textwidth}
	\centering
	\vspace{-5mm}
	\begin{subfigure}{\linewidth}
        \centering
        \scalebox{0.95}{\begin{tikzpicture}
	\draw[->] (-1.3, 0) -- (2.2, 0) node[right,scale=0.85] {$x$};
	\draw[->] (0, -0.4) -- (0, 2.2) node[above,scale=0.85] {$y$};

    \def\u{1.8}
    \def\l{-0.8}

    \coordinate (c) at (-1,0.8);
    \coordinate (d) at (0.2,0.3);
    \coordinate (e) at (1.3,1.0);
    \coordinate (f) at (2.0,2.1);
	
	\path[name path=F] (c) -- (d) -- (e) -- (f);
	\path[name path=Pl,overlay] ({\l},-2)--({\l},2.0);
	\path[name path=Pu,overlay] ({\u},-2)--({\u},2.0);
    \path[name path=Pd,overlay] ($(c)!2.cm!180:(d)$) -- ($(d)!2.cm!180:(c)$);
    \path[name path=Pe,overlay] ($(d)!2.cm!180:(e)$) -- ($(e)!2.cm!180:(d)$);
	
	\path [name intersections={of=Pl and Pd,by={cll}}];
	\path [name intersections={of=Pu and Pd,by={clu}}];
	\path [name intersections={of=Pl and F,by={cu}}];
	\path [name intersections={of=Pu and Pe,by={du}}];
	\path [name intersections={of=Pu and F,by={eu}}];
	\path [name intersections={of=Pl and Pe,by={ell}}];
	\coordinate (da) at ($(d)!2.cm!180:(e)$);
	\coordinate (db) at ($(e)!2.cm!180:(d)$);
	\coordinate (elu) at ($(da)!(eu)!(db)$);

	\fill[fill=black!90, pattern=north west lines, opacity=0.5] (cll) -- (eu) -- (clu) -- cycle;
	\fill[fill=green!85!black!90, opacity=0.35] (clu) -- (cll) -- (du) -- cycle;
	\fill[fill=red!90, opacity=0.35] (ell) -- (eu) -- (du) -- cycle;
	
	\draw[black,thick, name path=f] (c) -- (d) -- (e) -- (f);
    \draw[-, style=dashed] (d) -- ($(d)!0.9cm!180:(c)$);
    \draw[-, style=dashed, name path=fe] (e) -- ($(e)!0.9cm!180:(d)$);
	\draw[-] ({\l},-0.1) -- ({\l},0.1);
	\draw[-] ({\u},-0.1) -- ({\u},0.1);

	\node[circle, fill=black, minimum size=3pt,inner sep=0pt, outer sep=0pt] at (c) {};
	\node[circle, fill=green!85!black!80, minimum size=3.6pt,inner sep=0pt, outer sep=0pt] at (d) {};
	\node[circle, fill=red!75, minimum size=3.6pt,inner sep=0pt, outer sep=0pt] at (e) {};
	\node[circle, fill=black, minimum size=3pt,inner sep=0pt, outer sep=0pt] at (f) {};

	\node[anchor=north east,align=center,scale=0.85] at ({\l},-0.20) {$l$};
	\node[anchor=north west,align=center,scale=0.85] at ({\u},-0.20) {$u$};

\end{tikzpicture}}    
    \end{subfigure}
    \begin{subfigure}{\linewidth}
        \centering
        \scalebox{0.95}{\tikzset{
  half circle/.style={
      semicircle,
      shape border rotate=180,
      anchor=chord center,
	  line width=0
      }
}

\begin{tikzpicture}
	\draw[->] (-1.3, 0) -- (2.2, 0) node[right,scale=0.85] {$x$};
	\draw[->] (0, -0.4) -- (0, 2.2) node[above,scale=0.85] {$y$};

    \def\u{1.8}
    \def\l{-0.8}

    \coordinate (c) at (-1,0.8);
    \coordinate (d) at (0.2,0.3);
    \coordinate (e) at (1.3,1.0);
    \coordinate (f) at (2.0,2.1);

	\path[name path=F] (c) -- (d) -- (e) -- (f);

	\path[name path=Pl,overlay] ({\l},-1)--({\l},2.0);
	\path[name path=Pu,overlay] ({\u},-1)--({\u},2.0);
	\path[name path=Pd,overlay] ($(d) + (180:2)$) -- ($(d) + (0:2)$);
    \path[name path=Pe,overlay] ($(d)!2.cm!180:(e)$) -- ($(e)!2.cm!180:(d)$);
	
	\path [name intersections={of=Pl and Pd,by={cll}}];
	\path [name intersections={of=Pu and Pd,by={clu}}];
	\path [name intersections={of=Pl and F,by={cu}}];
	\path [name intersections={of=Pu and Pe,by={du}}];
	\path [name intersections={of=Pu and F,by={eu}}];
	\path [name intersections={of=Pl and Pe,by={ell}}];
	\coordinate (da) at ($(d)!2.cm!180:(e)$);
	\coordinate (db) at ($(e)!2.cm!180:(d)$);
	\coordinate (elu) at ($(da)!(eu)!(db)$);

	\fill[fill=black!90, pattern=north west lines, opacity=0.5] (cll) -- (cu) -- (eu) -- (clu) -- cycle;
	\fill[fill=blue!90, opacity=0.35] (clu) -- (cll) -- (cu) -- cycle;
	\fill[fill=green!85!black!90, opacity=0.35] (clu) -- (cll) -- (du) -- cycle;
	\fill[fill=red!90, opacity=0.35] (ell) -- (eu) -- (du) -- cycle;

	\draw[black,thick] (c) -- (d) -- (e) -- (f);
    \draw[-, style=dashed] (d) -- ++(180:0.9cm);
    \draw[-, style=dashed] (d) -- ++(0:0.9cm);
    \draw[-, style=dashed, name path=fe] (e) -- ($(e)!0.9cm!180:(d)$);
	\draw[-] ({\l},-0.1) -- ({\l},0.1);
	\draw[-] ({\u},-0.1) -- ({\u},0.1);

	\node[circle, fill=black, minimum size=3pt,inner sep=0pt, outer sep=0pt] at (c) {};
	\node[half circle, rotate=90, fill=green!85!black!80, minimum size=2pt,inner sep=0pt, outer sep=0pt] at (d) {};
	\node[half circle, rotate=270, fill=blue!80, minimum size=2pt,inner sep=0pt, outer sep=0pt] at (d) {};
	\node[circle, fill=red!75, minimum size=3.6pt,inner sep=0pt, outer sep=0pt] at (e) {};
	\node[circle, fill=black, minimum size=3pt,inner sep=0pt, outer sep=0pt] at (f) {};

	\node[anchor=north east,align=center,scale=0.85] at ({\l},-0.20) {$l$};
	\node[anchor=north west,align=center,scale=0.85] at ({\u},-0.20) {$u$};

\end{tikzpicture}}    
    \end{subfigure}
	\begin{subfigure}{\linewidth}
        \centering
        \scalebox{0.95}{\begin{tikzpicture}

	\draw[->] (-1.3, 0) -- (2.2, 0) node[right,scale=0.85] {$x$};
	\draw[->] (0, -0.4) -- (0, 2.2) node[above,scale=0.85] {$y$};

    \def\u{1.8}
    \def\l{-0.8}

    \coordinate (c) at (-1,0.8);
    \coordinate (d) at (0.2,0.3);
    \coordinate (e) at (1.3,1.0);
    \coordinate (f) at (2.0,2.1);
	
	\path[name path=F] (c) -- (d) -- (e) -- (f);
	\path[name path=Pl,overlay] ({\l},-2)--({\l},2.0);
	\path[name path=Pu,overlay] ({\u},-2)--({\u},2.0);
    \path[name path=Pd,overlay] ($(c)!2.cm!180:(d)$) -- ($(d)!2.cm!180:(c)$);
    \path[name path=Pe,overlay] ($(d)!2.cm!180:(e)$) -- ($(e)!2.cm!180:(d)$);
	
	\path [name intersections={of=Pl and Pd,by={cll}}];
	\path [name intersections={of=Pu and Pd,by={clu}}];
	\path [name intersections={of=Pl and F,by={cu}}];
	\path [name intersections={of=Pu and Pe,by={du}}];
	\path [name intersections={of=Pu and F,by={eu}}];
	\path [name intersections={of=Pl and Pe,by={ell}}];
	\coordinate (da) at ($(d)!2.cm!180:(e)$);
	\coordinate (db) at ($(e)!2.cm!180:(d)$);
	\coordinate (elu) at ($(da)!(eu)!(db)$);

	\fill[color=black!90, pattern=north west lines, opacity=0.5] (cll) -- (d) -- (e) -- (eu) -- cycle;
	\fill[fill=green!85!black!90, opacity=0.35] (d) -- (cll) -- (du) -- cycle;
	\fill[fill=red!90, opacity=0.35] (ell) -- (eu) -- (e) -- cycle;

	\draw[black,thick] (c) -- (d) -- (e) -- (f);
    \draw[-, style=dashed] (d) -- ($(d)!0.9cm!180:(c)$);
    \draw[-, style=dashed, name path=fe] (e) -- ($(e)!0.9cm!180:(d)$);
	\draw[-] ({\l},-0.1) -- ({\l},0.1);
	\draw[-] ({\u},-0.1) -- ({\u},0.1);
	
	\node[circle, fill=black, minimum size=3pt,inner sep=0pt, outer sep=0pt] at (c) {};
	\node[circle, fill=green!85!black!80, minimum size=3.6pt,inner sep=0pt, outer sep=0pt] at (d) {};
	\node[circle, fill=red!80, minimum size=3.6pt,inner sep=0pt, outer sep=0pt] at (e) {};
	\node[circle, fill=black, minimum size=3pt,inner sep=0pt, outer sep=0pt] at (f) {};

	\node[anchor=north east,align=center,scale=0.85] at ({\l},-0.20) {$l$};
	\node[anchor=north west,align=center,scale=0.85] at ({\u},-0.20) {$u$};

\end{tikzpicture}}    
    \end{subfigure}
	\vspace{-3.5mm}
	\caption{Illustration of two different ReLU network encodings of the same function unde \dpO- (top and middle) and \triangle-analysis (bottom).}
	\label{fig:DPO_convex}
\end{wrapfigure}
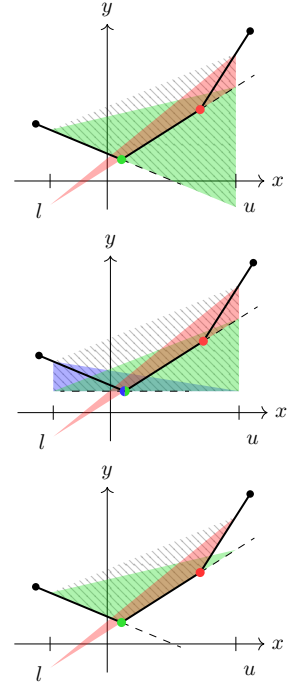
We show constructively that finite ReLU networks can \dpO-express C-CPWL functions, by first encoding any such function as a single-layer ReLU network. We note that the below results equivalently apply to concave functions:

\begin{restatable}[Convex encoding]{lem}{ConvModeling}\label{lem:conv_modeling}
    Let $f \in \cpwl(\I, \R)$ be convex. Then $f$ is encoded by
    \begin{equation}\label{eqn:single_layer_monotone}
        h(x) = b + c x + \sum_{i=1}^{n-1} \gamma_i \RR(\pm_i(x-x_i)),
    \end{equation}
    for any choice $\pm_i \in \{-1, 1\}$, if $b$ and $c$ are set appropriately, where $\alpha_i = \tfrac{f(x_{i+1}) - f(x_i)}{x_{i+1} - x_i}$ is the slope between points $x_i$ and $x_{i+1}$, and $\gamma_i = \alpha_i - \alpha_{i-1} > 0$ the slope change at $x_{i+1}$. 
\end{restatable}

Intuitively, we encode the C-CPWL function $f$ by starting with a linear function $h_0 = b + c x$, coinciding with one of the linear segments of $f$. We then pick one of the points $x_i$ where $f$ changes slope that are adjacent to this segment and add $\RR(\pm_i (x - x_i))$ changing its activation state at this point. Regardless of $\pm_i$, we now scale this ReLU with $\gamma_i = \alpha_i - \alpha_{i-1}$ to introduce the local change of slope, and update the linear term $c \gets c - \gamma_i$ if the newly added ReLU affects the segment that the linear function matched originally. We repeat this process until $h$ encodes $f$. 

We illustrate this in \cref{fig:DPO_convex} (top), where we start our construction with the left-most linear segment. We continue by adding a ReLU, first at the green and then the red point, and show the \dpO relaxation of the added ReLUs as a shaded area of the same color. 
We illustrate the resulting overall \dpO-relaxation, obtained as their point-wise sum, striped grey.
Observe that this always recovers the original linear term as the lower bound (see \cref{fig:DPO_convex} top). This leads to an imprecise output range unless its slope $c$ is 0. If $f$ includes such a constant section with zero-slope, we can directly apply the above construction, always changing $\pm_i$ such that the ReLUs open "outward", i.e., in a direction that does not affect the constant segment. If $f$ does not include such a constant section but a unique minimum, as in our example, we place two ReLUs at this point, treating it as a constant section with 0-width and recovering a precise lower bound (see \cref{fig:DPO_convex} middle).  Thus finite ReLU networks can \dpO-express C-CPWL functions but do not allow $\pm_i$ to be chosen freely. Note that the upper bound is still precise regardless of the choice of $\pm_i$.

\begin{restatable}[\dpO Convex]{thm}{DeeppolyConvex} \label{thm:deeppoly_convex}
    For any convex CPWL function $f \colon \I \to \R$, there exists exactly one network of the form $h(x) = b + \sum_{i \in \bc{I}} \gamma_i \RR(\pm_i(x-x_i))$ encoding $f$, with $|\bc{I}| = n-1$ if $f$ has slope zero on some segment and otherwise $|\bc{I}| = n$, such that its \dpO-analysis is precise, where $\gamma_i > 0$ for all $i$. 
\end{restatable}

\subsection{DeepPoly-1}

To show that \dpo has the same expressivity as \dpO, we encode a ReLU function as $h(x) = x + \RR(-x)$ which under \dpo-analysis yields the same linear bounds as $h'(x) = \RR(x)$ under \dpO-analysis. The reverse also holds. Thus, the expressivity of \dpO and \dpo is equivalent.

\begin{restatable}[\dpo ReLU]{cor}{DPoReLU}\label{thm:dp1_ReLU}
    The ReLU network $h(x) = x + \RR(-x)$ encodes the function $f(x) = \RR(x)$ and, the \dpo-analysis of $h(x)$ is identical to the \dpO-analysis of $\RR$. Further, the \dpO-analysis of $h(x)$ is identical to the \dpo-analysis of $\RR$.
\end{restatable}

It follows directly that any function that can be \dpO-expressed by a finite ReLU network can be \dpo-expressed by the same ReLU network after substituting every $\RR(x)$ with $x + \RR(-x)$:

\begin{restatable}[\dpo Approximation]{cor}{DPoClass}\label{thm:dp1_class}
    Finite $\RR$ networks can \dpo- and \dpO-express the same function class precisely. In particular, they can \dpo-express the set of convex functions $f \in \cpwl(\I, \R)$ and monotone functions $f \in \cpwl(\I, \R)$ precisely. 
\end{restatable}

\subsection{Triangle}

To show that finite ReLU networks can \triangle-express C-CPWL functions, we reuse the construction from \cref{lem:conv_modeling}. 
However, as the \triangle-relaxation yields the exact convex hull for ReLU functions, we first show that the convex hull of a sum of convex functions (such as \cref{eqn:single_layer_monotone}) is recovered by the pointwise sum of their convex hulls:

\begin{restatable}[Convex Hull Sum]{lem}{ConvexHullSums}\label{thm:convex_hull_sums}
    Given two convex functions $f, g\colon \R \to \R$ and the box $[l, u]$. Then, the pointwise sum of the convex hulls $\bc{H}_f + \bc{H}_g$ is identical to the convex hull of the sum of the two functions $\bc{H}_{f+g} = \bc{H}_f + \bc{H}_g$. 
\end{restatable}

This follows directly from the definition and implies that the \triangle-analysis is precise for arbitrary choices of $\pm_i$, illustrated in the bottom of \cref{fig:DPO_convex}:

\begin{restatable}[\triangle Convex]{thm}{TriangleConv}\label{lem:triangle_conv}
    Let $f \in \cpwl(\I, \R)$ be convex. Then, for any network $h$ encoding $f$ as in \cref{lem:conv_modeling}, we have that its \triangle-analysis is precise. In particular, $\pm_i$ can be chosen freely. 
\end{restatable}

\subsection{Multi-Neuron-Relaxations}

As multi-neuron relaxations yield the exact convex hull of the considered group of neurons (all within the same layer), it is sufficient to show that we can express arbitrary CPWL functions with a single-layer network to see that they \mn-express CPWL functions. To this end, we use a similar construction as in \cref{lem:conv_modeling}, where the lack of convexity removes the positivity constraint on $\gamma_i$. 

\begin{restatable}[Multi-Neuron Precision]{thm}{multi_neuron}\label{thm:multi_neuron}
    For every $f \in \cpwl(\I, \R)$, there exists a single layer ReLU network $h$ encoding $f$, such that its \mn-analysis (considering all ReLUs jointly) is precise.
\end{restatable}

\section{Convex Relaxations for Multivariate Functions}

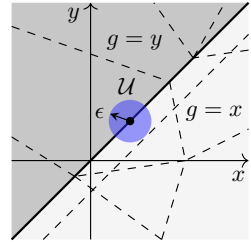
\begin{wrapfigure}[15]{r}{0.35\textwidth}
	\centering
	\vspace{-4mm}
	\scalebox{1.05}{\begin{tikzpicture}
	\draw[->] (-1., 0) -- (2., 0) node[yshift=-1.2ex, xshift=-0.8ex,scale=0.85] {$x$};
	\draw[->] (0, -1.0) -- (0, 2.0) node[yshift=-1.ex, xshift=-0.2ex,left,scale=0.85] {$y$};

    \coordinate (a) at (1.,1.);
    \coordinate (b) at (-.2,-.2);
    \coordinate (c) at (1.3,1.3);
    \coordinate (e) at (1.2,0);
    \coordinate (f) at (2.0, 0.4);
    \coordinate (g) at (2.0,1.4);
    \coordinate (h) at (1.7,2);
    \coordinate (i) at (0.8,2);
    \coordinate (j) at (0.9,-1);
    \coordinate (k) at (-1,.3);
    \coordinate (l) at (-1,1.7);
    \coordinate (m) at (-0.6,-1);
    \coordinate (n) at (2,1.6);

    \coordinate (d) at (0.5,0.5);
		
    \draw[black,thick] (-1, -1) -- (2, 2);
    \fill[black, opacity=0.04] (-1, -1) -- (2, 2) -- (2,-1) -- cycle;
    \fill[black, opacity=0.22] (-1, -1) -- (2, 2) -- (-1,2) -- cycle;

    \node[circle, fill=blue!80, minimum size=5.4mm,inner sep=0pt, outer sep=0pt, opacity=0.5] at (d) {};

    \node[circle, fill=black, minimum size=3pt,inner sep=0pt, outer sep=0pt] at (d) {};
    \draw[-stealth] (d) -- ++(160:2.7mm);

    \node[anchor=west,align=center,scale=0.85] at ($(d)+(-0.55,.1)$) {$\epsilon$};
    \node[anchor=west,align=center,scale=0.85] at ($(d)+(-0.25,.45)$) {$\bc{U}$};
	
	\draw[-, style=dashed] (a) -- (e) -- (f);
    \draw[-, style=dashed] (g) -- (c);
    \draw[-, style=dashed] (e) -- (b);
    \draw[-, style=dashed] (c) -- (h);
    \draw[-, style=dashed] (c) -- (i);
    \draw[-, style=dashed] (j) -- (e);
    \draw[-, style=dashed] (j) -- (b);
    \draw[-, style=dashed] (k) -- (b);
    \draw[-, style=dashed] (l) -- (a);
    \draw[-, style=dashed] (m) -- (n);
	\node[anchor=north west,align=center,scale=0.85] at (1.1, 0.80) {$g = x$};
	\node[anchor=north west,align=center,scale=0.85] at (0.1, 1.7) {$g = y$};

\end{tikzpicture}}
	\vspace{-1mm}
	\caption{Illustration of the pre-image of activation pattern changes with (\acfc) and without (\acnfc) functional change of a ReLU network $h$ encoding the $g = \max(x,y)$ function, as well as the $\epsilon$-neighborhood $\bc{U}$ (\neigh), in which the only activation change occurs at $x=y$.}
	\label{fig:locality}
\end{wrapfigure}
After having shown in the previous section that, for univariate functions, the expressivity of ReLU networks under convex relaxations is not fundamentally limited, we now turn our attention to multivariate functions. 
There, we prove that no finite ReLU network can \triangle-express the maximum function $\max \colon \R^2 \to \R$ precisely (\cref{thm:triangle_max}). This directly implies that no single-neuron relaxation can express the class of multivariate, monotone, and convex CPWL functions precisely.%

Intuitively, we will argue along the following lines. We first observe that for any finite ReLU-Network $h$ that encodes the maximum function, we can find a point $(x, y=x) \in \R^2$ with neighborhood $\bc{U}$, such that on $\bc{U}$, all ReLUs in $h$ either switch their activation state for $x = y$ or not at all (see \cref{fig:locality}).
Then, we show that for such a neighborhood, we can \triangle-replace the finite ReLU network $h$ with a single layer consisting of just 2 neurons and a linear term (\cref{thm:network_form_coverage,thm:network_simplification}). Finally, we show that no such single-layer network can \triangle-express $\max$ precisely (\cref{thm:triangle_single_max}), before putting everything together in \cref{thm:triangle_max}. All proofs and support lemmata are again deferred to \cref{app:theory_multi}.

Let us begin by showing that we can express \emph{any} finite ReLU network using the functional form of \cref{eqn:network_form}. 
That is, every $i$-layer network $\vh^i$ can be written as the sum of an ($i-1$)-layer network $\vh_L^{i-1}$ and a linear function of a ReLU applied to another ($i-1$)-layer network $\mW_i \RR(\vh_R^{i-1})$. 
Further, if, for a given input region $\bc{U}$, all ReLUs in the original network switch activation state on the hyperplane $\vw^\top \vx=0$ or not at all, then, we can ensure that \emph{every} ReLU in both ($i-1$)-layer networks change activation state exactly for $z \coloneqq \vw^\top \vx = 0$.

\begin{restatable}[Network Form Coverage]{thm}{NetworkFormCoverage}\label{thm:network_form_coverage}
    Given a neighborhood $\bc{U}$ and a finite $k$-layer ReLU network $h$ such that on $\bc{U}$ and under \triangle-analysis all its ReLUs are either stably active ($\RR(v) = v$), stably inactive ($\RR(v) = 0$), or switch activation state for $z \coloneqq \vw^\top \vx = 0$ with $\vw \in \R^d$, then $h$ can be represented using the functional form 
    \begin{equation}
        \label{eqn:network_form}
        \vh_{\{R,L\}}^i = \vh_L^{i-1} + \mW_i \RR(\vh_R^{i-1}), \quad \vh_{\{R,L\}}^0 = \vb + \mW_0 \vx,
    \end{equation}
    for $i=k$ and such that all ReLUs switch their activation state at $\{\vx \in \bc{X} \mid \vw^\top \vx = 0\}$. Here, $L$ and $R$ are labels, used to distinguish the two possibly different $\vh^{i-1}$ from each other.
\end{restatable}

We can now leverage the fact that all ReLUs change activation state at the same point to simplify the linear terms of ReLUs to a sum of just two: $\sum_i a_i \RR(w_i z) \overset{\Delta}{\rightsquigarrow} \gamma z + \alpha \RR(z)$ for some $\gamma, \alpha \in \R$ (\cref{thm:triangle_sum}). This allows us to further simplify a ReLU applied to such a sum of ReLUs: $\RR(\gamma + \alpha \RR(z)) \overset{\Delta}{\rightsquigarrow} \gamma' z + \alpha' \RR(z)$ (\cref{thm:triangle_comp}). These two replacements allow us to recursively reduce the depth of networks in the form of \cref{eqn:network_form} until just a single layer is left: 

\begin{restatable}[Network Simplification]{thm}{NetworkSimplification}\label{thm:network_simplification}
    Let $h^k$ be a network as in \cref{thm:network_form_coverage} such that all ReLUs change activation state at $z \coloneqq \vw^\top \vx = 0$ with $\vw \in \R^d$. 
    We have
    \begin{equation*}
        h^k= h_L^{k-1} + \mW \RR(\vh_R^{k-1})
        \quad \overset{\Delta}{\rightsquigarrow} \quad
        h(\vx) = b + \mW \vx + \alpha \RR(z),
    \end{equation*}
    where $h^0(\vx) = b_0 + \mW_0 \vx$ and all $\RR$ change state exactly at $\{\vx \in \bc{X} \mid \vw^\top \vx = 0\}$.
\end{restatable}

\begin{figure}[t]
	\centering
	\vspace{-2mm}
    \begin{subfigure}{0.32\linewidth}
        \centering
        \scalebox{0.95}{\begin{tikzpicture}[rotate around x=-90,rotate around z=55]
    \def\s{1.8}
    \def\a{1.0}
    \def\d{10}
    
    \coordinate (O) at (-0,0,0);
    \coordinate (SW) at (0,0,0);
    \coordinate (SE) at ({\s},0,{\s});
    \coordinate (NE) at ({\s},{\s},{\s});
    \coordinate (NW) at (0,{\s},{\s});

    \coordinate (ONE) at ({\s},{\s},0);

    \coordinate (XNW) at (0,{\s},0);
    \coordinate (YSE) at ({\s},0,0);

    \coordinate (TSW) at (0,0,{0.5*\s});
    \coordinate (TNE) at ({\s},{\s},{1.5*\s});

    \draw[style=dashed] (ONE) -- (NE);
    \fill[color=black!80, opacity=0.10] (SW) -- (XNW) -- (ONE) -- (YSE) -- cycle;

    \fill[color=green!50, opacity=0.8] (SW) -- (SE) -- (NE) -- cycle;
    \fill[color=green!80!black!50, opacity=0.8] (SW) -- (NW) -- (NE) -- cycle;
    \draw[color=black!80, line width = 0.3pt] (SW) -- (SE) -- (NE);
    \draw[color=black!80, line width = 0.3pt] (SW) -- (NW) -- (NE) -- cycle;

    \draw[style=dashed] (XNW) -- (NW);
    \draw[style=dashed] (YSE) -- (SE);

    \draw ($(NE)!0.0!(SE)+(+0.,0, 0.0)$) node[anchor=south, scale=0.8] {$z = \max(x,y)$};

    \draw[->] (O) -- ($(O)+({1.1*\a},0,0)$);
    \draw[->] (O) -- ($(O)+(0,{1.1*\a},0)$);
    \draw[->] (O) -- ($(O)+(0,0,{1.1*\a})$);
    \draw ($(O)+(0,0,{1.1*\a})$) node[anchor=east, scale=0.9] {$z$};
    \draw ($(O)+({1.1*\a},0,0)$) node[anchor=north west, scale=0.9] {$x$};
    \draw ($(O)+(0,{1.1*\a},0)$) node[anchor=north east, scale=0.9] {$y$};

\end{tikzpicture}}    
        \hspace{-12mm}
    \end{subfigure}
    \hfil
    \begin{subfigure}{0.32\linewidth}
        \centering
        \scalebox{0.95}{\begin{tikzpicture}[rotate around x=-90,rotate around z=55]
    \def\s{1.8}
    \def\a{1.0}
    
    \coordinate (O) at (-0,0,0);
    \coordinate (SW) at (0,0,0);
    \coordinate (SE) at ({\s},0,{\s});
    \coordinate (NE) at ({\s},{\s},{\s});
    \coordinate (NW) at (0,{\s},{\s});

    \coordinate (XNW) at (0,{\s},0);
    \coordinate (YSE) at ({\s},0,0);

    \coordinate (TSW) at (0,0,{0.5*\s});
    \coordinate (TNE) at ({\s},{\s},{1.5*\s});

    \fill[color=black!80, opacity=0.10] (SW) -- (XNW) -- (ONE) -- (YSE) -- cycle;
    \draw[style=dashed] (ONE) -- (NE);

    \draw[color=black, line width = 0.3pt] (NE) -- (XNW) -- (SW);
    \draw[color=black, line width = 0.3pt] (NE) -- (YSE) -- (SW);
    \draw[line width = 0.3pt] (NE) -- (SW);
    \fill[color=black!35, opacity=0.7] (XNW) -- (SW) -- (SE) -- (NE)-- cycle;
    \fill[color=black!45, opacity=0.7] (NW) -- (SW) -- (YSE) -- (NE)-- cycle;
    \draw[color=black, line width = 0.3pt] (NE) -- (SE) -- (SW);
    \draw[color=black, line width = 0.3pt] (NE) -- (NW) -- (SW);

    \draw[style=dashed] (NW) -- (XNW);
    \draw[style=dashed] (SE) -- (YSE);

    \draw[line width = 0.3pt] (NE) -- (SW);

    \draw ($(NW)+(-0.05,0.0,0.3)$) node[anchor=south west, scale=0.8] {$z \geq x$};
    \draw ($(NE)!0.3!(SE)+(+0.2,0, 0.0)$) node[anchor= west, scale=0.8] {$z \geq y$};

    \phantom{
    \draw[->] (O) -- ($(O)+({1.1*\a},0,0)$);
    \draw[->] (O) -- ($(O)+(0,{1.1*\a},0)$);
    \draw[->] (O) -- ($(O)+(0,0,{1.1*\a})$);
    \draw ($(O)+(0,0,{1.1*\a})$) node[anchor=east, scale=0.9] {$z$};
    \draw ($(O)+({1.1*\a},0,0)$) node[anchor=north west, scale=0.9] {$x$};
    \draw ($(O)+(0,{1.1*\a},0)$) node[anchor=north east, scale=0.9] {$y$};
    }

\end{tikzpicture}}    
    \end{subfigure}
    \hfil
    \begin{subfigure}{0.32\linewidth}
        \centering
        \scalebox{0.95}{\begin{tikzpicture}[rotate around x=-90,rotate around z=55]
    \def\s{1.8}
    \def\a{1.0}
    
    \coordinate (O) at (-0,0,0);
    \coordinate (SW) at (0,0,0);
    \coordinate (SE) at ({\s},0,{\s});
    \coordinate (NE) at ({\s},{\s},{\s});
    \coordinate (NW) at (0,{\s},{\s});

    \coordinate (XNW) at (0,{\s},0);
    \coordinate (YSE) at ({\s},0,0);

    \coordinate (TSW) at (0,0,{0.5*\s});
    \coordinate (TNE) at ({\s},{\s},{1.5*\s});

    \draw[style=dashed] (ONE) -- (NE);

    \fill[color=black!80, opacity=0.10] (SW) -- (XNW) -- (ONE) -- (YSE) -- cycle;

    \fill[color=green!50, opacity=0.8] (SW) -- (SE) -- (NE) -- cycle;
    \fill[color=green!80!black!50, opacity=0.8] (SW) -- (NW) -- (NE) -- cycle;
    \draw[color=black!80, line width = 0.3pt] (SW) -- (SE) -- (NE);
    \draw[color=black!80, line width = 0.3pt] (SW) -- (NW) -- (NE) -- cycle;

    \draw[style=dashed] (TNE) -- (NE);
    \draw[style=dashed] (NW) -- (XNW);
    \draw[style=dashed] (SE) -- (YSE);

    \draw[style=dashed] (TSW) -- (SW);

    \draw[line width = 0.3pt] (NE) -- (SW);

    \phantom{\draw ($(SE)!0.6!(TNE)+(0.1,0)$) node[anchor=west, scale=0.8] {$z \leq\tfrac{x+y+1}{2}$};}
    \phantom{\draw ($(YSE)+(+0.05,0, 0.3)$) node[anchor=west, scale=0.8] {$z \geq y$};}
    \phantom{\draw ($(SW)!0.85!(SE)+(+0.2,0, 0.0)$) node[anchor=west, scale=0.8] {$z = \max(x,y)$};}

    \draw ($(SE)!0.6!(TNE)+(0.1,0)$) node[anchor=west, scale=0.8] {$z \leq\tfrac{x+y+1}{2}$};

    \fill[color=blue!60, opacity=0.7] (TSW) -- (SE) -- (TNE) -- (NW) -- cycle;
    \draw[color=black, line width = 0.3pt] (TSW) -- (SE) -- (TNE) -- (NW) -- cycle;

    \phantom{
    \draw[->] (O) -- ($(O)+({1.1*\a},0,0)$);
    \draw[->] (O) -- ($(O)+(0,{1.1*\a},0)$);
    \draw[->] (O) -- ($(O)+(0,0,{1.1*\a})$);
    \draw ($(O)+(0,0,{1.1*\a})$) node[anchor=east, scale=0.9] {$z$};
    \draw ($(O)+({1.1*\a},0,0)$) node[anchor=north west, scale=0.9] {$x$};
    \draw ($(O)+(0,{1.1*\a},0)$) node[anchor=north east, scale=0.9] {$y$};
    }

\end{tikzpicture}}    
    \end{subfigure}
	\vspace{-2mm}
	\caption{Illustration of $z = \max(x,y)$ (left) with its \triangle lower (middle) and upper (right) bounds as obtained in \cref{thm:triangle_single_max}.}
	\label{fig:max}
	\vspace{-3mm}
\end{figure}
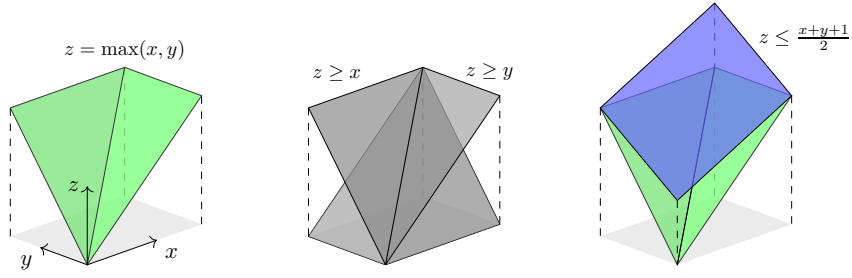

Note that, $h^k$, $h_L^{k-1}$ and $h$ map to $\R$, while $\vh^{k-1}_R$ maps to the space of some hidden layer $\R^n$. Next, we show directly via contradiction that single-layer ReLU networks of this form cannot \triangle-express the maximum function, illustrating the resulting (imprecise) bounds in \cref{fig:max}:

\begin{restatable}[Triangle $\max$]{thm}{TriangleSingleMax} \label{thm:triangle_single_max}
    $\RR$-networks of the form $h(x,y) = b + w_x x + w_y y + \alpha \relu(x-y)$ can not \triangle-express the function $\max \colon \R^2 \to \R$. 
\end{restatable}
\begin{Proof}
    We consider the input region $\bc{B} = [0,1]^2$ and constrain our parameters by considering the following: For $x = y = 0$, we have $f(0,0) = 0 = b = h(0,0)$. For $x < y$, we have $f(x,y) = y = w_x x + w_y y = h(x,y)$ and thus $w_x = 0$, $w_y = 1$. Finally for $x>y$, we have $f(x,y) = x = y + \alpha (x-y) = h(x,y)$ and thus $\alpha = 1$.
    Hence we have $h(x,y) = y + \RR(x-y)$. 
    \begin{align*}
        \begin{rcases}
            0 \\
            x-y
        \end{rcases}
        \leq
        \RR(x-y)
        \leq
        \tfrac{1}{2}(x - y + 1)
        \quad
        \implies
        \quad
        \begin{rcases}
            y \\
            x
        \end{rcases}
        \leq
        h(x,y)
        \leq
        \tfrac{1}{2}(x + y + 1). 
    \end{align*}
    The maximum of the upper bound is attained at $x = y = 1$, where we get $\overline{h^\Delta(\bc{B})} = \tfrac{3}{2}$ which is larger than $\overline{\max(\bc{B})} = 1$ (see \cref{fig:max}). 
\end{Proof}
To show that no ReLU network can \triangle-express $\max$ it remains to argue how we can find a $\bc{U}$ such that all ReLUs switch their activation state at $\{\vx \in \bc{X} \mid \vw^\top \vx = 0\}$ or not at all:

\begin{restatable}[$\Delta$ Impossibility $\max$]{thm}{TriangleMax}\label{thm:triangle_max}
    Finite $\RR$ networks can not \triangle-express the function $\max$. 
\end{restatable}

\begin{Proof}
We will prove this theorem via contradiction in four steps. Assume there exists a finite ReLU network $h$ that \triangle-expresses $\max$ precisely.

\textbf{First -- Locality} We argue this point in three steps:
    \vspace{-3mm}
    \begin{enumerate}
        \itemsep0em 
        \item There exists a point $(x,y=x)$ with an $\epsilon$-neighborhood $\bc{U}'$ such that one of the following holds for any $\RR(v)$ with input $v = h_v(x,y)$ of the network $h$:
            \begin{itemize}
                \item the ReLU is always active, i.e., $\forall (x,y) \in U, \RR(v) = v$,
                \item the ReLU is never active, i.e., $\forall (x,y) \in U, \RR(v) = 0$, or
                \item the ReLU changes activation state for $x=y$, i.e., $\exists v' \in \R, s.t., \RR(v) = \RR(v' (x-y))$.
            \end{itemize}
            This follows directly from the fact that finite ReLU networks divide their input space into finitely many linear regions (see the illustration in \cref{fig:locality}).
        \item   Further, there exists an $\epsilon$-neighborhood $\bc{U}$ of $(x,y)$ such that the above holds under \boxd-analysis, as it depends continuously on the input bounds and becomes exact for any network when the input bounds describe a point.
        \item Via rescaling and translation, we can assume that the point $(x,y)$ is at $\mathbf{0}$ and that the neighborhood $\bc{U}$ covers $[-1,1]^2$. 
    \end{enumerate}

\textbf{Second -- Network Form} 
On the neighborhood $\bc{U}$, any finite ReLU-network $h$ can, w.r.t. \triangle, be represented by $h^k = h^{k-1} + \mW \RR(\vh^{k-1}), h^0(v) = \vb + \mW \vv$ with biases $\vb \in \R^{d_k}$, weight matrices $\mW \in \R^{d_k \times d_{k-1}}$, where all ReLUs change activation state exactly for $x=y$ (\cref{thm:network_form_coverage}).

\textbf{Third -- Network Replacements}
We can replace $h^k$ w.r.t. \triangle with the single layer network $h'(x, y) = b + \mW (x,y)^\top + \alpha R(x-y)$ (\cref{thm:network_simplification}).

\textbf{Fourth -- Conclusion} There exists no network of this form encoding the $\max$-function such that its \triangle-analysis is precise on the interval $[0,1]^2$ (\cref{thm:triangle_single_max}).

This concludes the proof. 
\end{Proof}

As $\max$ belongs to the class of multivariate, convex, monotone, CPWL functions, it follows directly from \cref{thm:triangle_max} that no finite ReLU network can \triangle-express this class precisely:

\begin{restatable}[$\Delta$ Impossibility]{cor}{TriangleImpossibility} \label{cor:triangle_impossibility}
    Finite ReLU networks can not \triangle-express the set of convex, monotone, CPWL functions mapping from some box $\I \subset \R^2$ to $\R$. 
\end{restatable}

\section{Conclusion}
We conduct the first in-depth study on the expressivity of ReLU networks under all commonly used convex relaxations and find that: 
(i) more precise relaxations (\triangle, \dpO or \dpo) allow a larger class of \emph{univariate} functions (C-CPWL and M-CPWL) to be expressed precisely than the simple \boxd-relaxation (M-CPWL), 
(ii) for the same function class (C-CPWL), a more precise relaxation (\triangle vs \dpO or \dpo), can allow an exponentially larger solution space of ReLU networks,
(iii) \mn-relaxations allow single-layer networks to express all univariate CPWL functions,
(iv) even the most precise single-neuron relaxation (\triangle) is too imprecise to express \emph{multivariate}, convex, monotone CPWL functions precisely with finite ReLU networks, despite their exact analysis being trivial.

While more precise domains improve expressivity for univariate functions, all single-neuron convex-relaxations are fundamentally limited in the multivariate setting. 
Surprisingly, even simple functions that can be encoded with a single neuron $h = y + \RR(x-y) = \max(x,y)$, can not be \triangle-expressed precisely using any finite ReLU network. 
This highlights not only the importance of recent, more precise multi-neuron- and BaB-based neural network certification methods but also suggests more precise methods might be needed for training.

\message{^^JLASTBODYPAGE \thepage^^J}

\clearpage
\bibliography{references}

\begin{thebibliography}{38}
\providecommand{\natexlab}[1]{#1}
\providecommand{\url}[1]{\texttt{#1}}
\expandafter\ifx\csname urlstyle\endcsname\relax
  \providecommand{\doi}[1]{doi: #1}\else
  \providecommand{\doi}{doi: \begingroup \urlstyle{rm}\Url}\fi

\bibitem[Baader et~al.(2020)Baader, Mirman, and Vechev]{BaaderMV20}
Maximilian Baader, Matthew Mirman, and Martin~T. Vechev.
\newblock Universal approximation with certified networks.
\newblock In \emph{Proc. of ICLR}, 2020.

\bibitem[Balunovic \& Vechev(2020)Balunovic and Vechev]{BalunovicV20}
Mislav Balunovic and Martin~T. Vechev.
\newblock Adversarial training and provable defenses: Bridging the gap.
\newblock In \emph{Proc. of ICLR}, 2020.

\bibitem[Biggio et~al.(2013)Biggio, Corona, Maiorca, Nelson, Srndic, Laskov,
  Giacinto, and Roli]{BiggioCMNSLGR13}
Battista Biggio, Igino Corona, Davide Maiorca, Blaine Nelson, Nedim Srndic,
  Pavel Laskov, Giorgio Giacinto, and Fabio Roli.
\newblock Evasion attacks against machine learning at test time.
\newblock In \emph{Proc of ECML PKDD}, 2013.
\newblock \doi{10.1007/978-3-642-40994-3\_25}.

\bibitem[Bunel et~al.(2020)Bunel, Lu, Turkaslan, Torr, Kohli, and
  Kumar]{BunelLTTKK20}
Rudy Bunel, Jingyue Lu, Ilker Turkaslan, Philip H.~S. Torr, Pushmeet Kohli, and
  M.~Pawan Kumar.
\newblock Branch and bound for piecewise linear neural network verification.
\newblock \emph{J. Mach. Learn. Res.}, 2020.

\bibitem[Dathathri et~al.(2020)Dathathri, Dvijotham, Kurakin, Raghunathan,
  Uesato, Bunel, Shankar, Steinhardt, Goodfellow, Liang, and
  Kohli]{DathathriDKRUBS20}
Sumanth Dathathri, Krishnamurthy Dvijotham, Alexey Kurakin, Aditi Raghunathan,
  Jonathan Uesato, Rudy Bunel, Shreya Shankar, Jacob Steinhardt, Ian~J.
  Goodfellow, Percy Liang, and Pushmeet Kohli.
\newblock Enabling certification of verification-agnostic networks via
  memory-efficient semidefinite programming.
\newblock In \emph{Proc. of NeurIPS}, 2020.

\bibitem[Dutta et~al.(2018)Dutta, Jha, Sankaranarayanan, and
  Tiwari]{DuttaJST18}
Souradeep Dutta, Susmit Jha, Sriram Sankaranarayanan, and Ashish Tiwari.
\newblock Output range analysis for deep feedforward neural networks.
\newblock In \emph{Proc. of NFM}, 2018.
\newblock \doi{10.1007/978-3-319-77935-5\_9}.

\bibitem[Dvijotham et~al.(2018)Dvijotham, Stanforth, Gowal, Mann, and
  Kohli]{DvijothamSGMK18}
Krishnamurthy Dvijotham, Robert Stanforth, Sven Gowal, Timothy~A. Mann, and
  Pushmeet Kohli.
\newblock A dual approach to scalable verification of deep networks.
\newblock In \emph{Proc. of UAI}, 2018.

\bibitem[Ehlers(2017)]{Ehlers17}
R{\"{u}}diger Ehlers.
\newblock Formal verification of piece-wise linear feed-forward neural
  networks.
\newblock In \emph{ATVA}, 2017.
\newblock \doi{10.1007/978-3-319-68167-2\_19}.

\bibitem[Ferrari et~al.(2022)Ferrari, M{\"{u}}ller, Jovanovic, and
  Vechev]{FerrariMJV22}
Claudio Ferrari, Mark~Niklas M{\"{u}}ller, Nikola Jovanovic, and Martin~T.
  Vechev.
\newblock Complete verification via multi-neuron relaxation guided
  branch-and-bound.
\newblock In \emph{Proc. of ICLR}, 2022.

\bibitem[Gehr et~al.(2018)Gehr, Mirman, Drachsler{-}Cohen, Tsankov, Chaudhuri,
  and Vechev]{GehrMDTCV18}
Timon Gehr, Matthew Mirman, Dana Drachsler{-}Cohen, Petar Tsankov, Swarat
  Chaudhuri, and Martin~T. Vechev.
\newblock {AI2:} safety and robustness certification of neural networks with
  abstract interpretation.
\newblock In \emph{S\&P}, 2018.
\newblock \doi{10.1109/SP.2018.00058}.

\bibitem[Gowal et~al.(2018)Gowal, Dvijotham, Stanforth, Bunel, Qin, Uesato,
  Arandjelovic, Mann, and Kohli]{GowalIBP2018}
Sven Gowal, Krishnamurthy Dvijotham, Robert Stanforth, Rudy Bunel, Chongli Qin,
  Jonathan Uesato, Relja Arandjelovic, Timothy~A. Mann, and Pushmeet Kohli.
\newblock On the effectiveness of interval bound propagation for training
  verifiably robust models.
\newblock \emph{ArXiv preprint}, abs/1810.12715, 2018.

\bibitem[He et~al.(2020)He, Li, Xu, and Zheng]{HeLXZ20}
Juncai He, Lin Li, Jinchao Xu, and Chunyue Zheng.
\newblock Relu deep neural networks and linear finite elements.
\newblock \emph{Journal of Computational Mathematics}, \penalty0 (3), 2020.
\newblock ISSN 1991-7139.
\newblock \doi{https://doi.org/10.4208/jcm.1901-m2018-0160}.

\bibitem[Jovanovic et~al.(2022)Jovanovic, Balunovic, Baader, and
  Vechev]{JovanovicBBV22}
Nikola Jovanovic, Mislav Balunovic, Maximilian Baader, and Martin~T. Vechev.
\newblock On the paradox of certified training.
\newblock \emph{Trans. Mach. Learn. Res.}, 2022.

\bibitem[Katz et~al.(2017)Katz, Barrett, Dill, Julian, and
  Kochenderfer]{KatzBDJK17}
Guy Katz, Clark~W. Barrett, David~L. Dill, Kyle Julian, and Mykel~J.
  Kochenderfer.
\newblock Reluplex: An efficient {SMT} solver for verifying deep neural
  networks.
\newblock \emph{ArXiv preprint}, abs/1702.01135, 2017.

\bibitem[Mao et~al.(2023)Mao, M{\"{u}}ller, Fischer, and Vechev]{MaoMFV23}
Yuhao Mao, Mark~Niklas M{\"{u}}ller, Marc Fischer, and Martin~T. Vechev.
\newblock {TAPS:} connecting certified and adversarial training.
\newblock In \emph{Proc. of NeurIPS}, 2023.

\bibitem[Mirman et~al.(2018)Mirman, Gehr, and Vechev]{MirmanGV18}
Matthew Mirman, Timon Gehr, and Martin~T. Vechev.
\newblock Differentiable abstract interpretation for provably robust neural
  networks.
\newblock In \emph{Proc. of ICML}, 2018.

\bibitem[Mirman et~al.(2022)Mirman, Baader, and Vechev]{MirmanBV22}
Matthew Mirman, Maximilian Baader, and Martin~T. Vechev.
\newblock The fundamental limits of neural networks for interval certified
  robustness.
\newblock \emph{Trans. Mach. Learn. Res.}, 2022.

\bibitem[M{\"{u}}ller et~al.(2022)M{\"{u}}ller, Makarchuk, Singh,
  P{\"{u}}schel, and Vechev]{MullerMSPV22}
Mark~Niklas M{\"{u}}ller, Gleb Makarchuk, Gagandeep Singh, Markus
  P{\"{u}}schel, and Martin~T. Vechev.
\newblock {PRIMA:} general and precise neural network certification via
  scalable convex hull approximations.
\newblock In \emph{Proc. of POPL}, 2022.
\newblock \doi{10.1145/3498704}.

\bibitem[M{\"{u}}ller et~al.(2023)M{\"{u}}ller, Eckert, Fischer, and
  Vechev]{MullerE0V23}
Mark~Niklas M{\"{u}}ller, Franziska Eckert, Marc Fischer, and Martin~T. Vechev.
\newblock Certified training: Small boxes are all you need.
\newblock In \emph{Proc. of ICLR}, 2023.

\bibitem[Palma et~al.(2021)Palma, Behl, Bunel, Torr, and Kumar]{PalmaBBTK21}
Alessandro~De Palma, Harkirat~S. Behl, Rudy Bunel, Philip H.~S. Torr, and
  M.~Pawan Kumar.
\newblock Scaling the convex barrier with active sets.
\newblock In \emph{Proc. of ICLR}, 2021.

\bibitem[Palma et~al.(2023)Palma, Bunel, Dvijotham, Kumar, Stanforth, and
  Lomuscio]{PalmaBDKSL23}
Alessandro~De Palma, Rudy Bunel, Krishnamurthy Dvijotham, M.~Pawan Kumar,
  Robert Stanforth, and Alessio Lomuscio.
\newblock Expressive losses for verified robustness via convex combinations.
\newblock \emph{CoRR}, abs/2305.13991, 2023.
\newblock \doi{10.48550/arXiv.2305.13991}.

\bibitem[Qin et~al.(2019)Qin, Dvijotham, O'Donoghue, Bunel, Stanforth, Gowal,
  Uesato, Swirszcz, and Kohli]{QinDOBSGUSK19}
Chongli Qin, Krishnamurthy~(Dj) Dvijotham, Brendan O'Donoghue, Rudy Bunel,
  Robert Stanforth, Sven Gowal, Jonathan Uesato, Grzegorz Swirszcz, and
  Pushmeet Kohli.
\newblock Verification of non-linear specifications for neural networks.
\newblock In \emph{Proc. of ICLR}, 2019.

\bibitem[Raghunathan et~al.(2018)Raghunathan, Steinhardt, and
  Liang]{RaghunathanSL18}
Aditi Raghunathan, Jacob Steinhardt, and Percy Liang.
\newblock Semidefinite relaxations for certifying robustness to adversarial
  examples.
\newblock In \emph{Proc. of NeurIPS}, 2018.

\bibitem[Salman et~al.(2019)Salman, Yang, Zhang, Hsieh, and
  Zhang]{SalmanY0HZ19}
Hadi Salman, Greg Yang, Huan Zhang, Cho{-}Jui Hsieh, and Pengchuan Zhang.
\newblock A convex relaxation barrier to tight robustness verification of
  neural networks.
\newblock In \emph{Proc. of NeurIPS}, 2019.

\bibitem[Singh et~al.(2018)Singh, Gehr, Mirman, P{\"{u}}schel, and
  Vechev]{SinghGMPV18}
Gagandeep Singh, Timon Gehr, Matthew Mirman, Markus P{\"{u}}schel, and
  Martin~T. Vechev.
\newblock Fast and effective robustness certification.
\newblock In \emph{Proc. of NeurIPS}, 2018.

\bibitem[Singh et~al.(2019{\natexlab{a}})Singh, Ganvir, P{\"{u}}schel, and
  Vechev]{SinghGPV19B}
Gagandeep Singh, Rupanshu Ganvir, Markus P{\"{u}}schel, and Martin~T. Vechev.
\newblock Beyond the single neuron convex barrier for neural network
  certification.
\newblock In \emph{Proc. of NeurIPS}, 2019{\natexlab{a}}.

\bibitem[Singh et~al.(2019{\natexlab{b}})Singh, Gehr, P{\"{u}}schel, and
  Vechev]{SinghGPV19}
Gagandeep Singh, Timon Gehr, Markus P{\"{u}}schel, and Martin~T. Vechev.
\newblock An abstract domain for certifying neural networks.
\newblock \emph{Proc. of POPL}, 2019{\natexlab{b}}.
\newblock \doi{10.1145/3290354}.

\bibitem[Szegedy et~al.(2014)Szegedy, Zaremba, Sutskever, Bruna, Erhan,
  Goodfellow, and Fergus]{SzegedyZSBEGF13}
Christian Szegedy, Wojciech Zaremba, Ilya Sutskever, Joan Bruna, Dumitru Erhan,
  Ian~J. Goodfellow, and Rob Fergus.
\newblock Intriguing properties of neural networks.
\newblock In \emph{Proc. of ICLR}, 2014.

\bibitem[Tjandraatmadja et~al.(2020)Tjandraatmadja, Anderson, Huchette, Ma,
  Patel, and Vielma]{TjandraatmadjaA20}
Christian Tjandraatmadja, Ross Anderson, Joey Huchette, Will Ma, Krunal Patel,
  and Juan~Pablo Vielma.
\newblock The convex relaxation barrier, revisited: Tightened single-neuron
  relaxations for neural network verification.
\newblock In \emph{Proc. of NeurIPS}, 2020.

\bibitem[Tjeng et~al.(2019)Tjeng, Xiao, and Tedrake]{TjengXT19}
Vincent Tjeng, Kai~Y. Xiao, and Russ Tedrake.
\newblock Evaluating robustness of neural networks with mixed integer
  programming.
\newblock In \emph{Proc. of ICLR}, 2019.

\bibitem[Wang et~al.(2021)Wang, Zhang, Xu, Lin, Jana, Hsieh, and
  Kolter]{WangZXLJHK21}
Shiqi Wang, Huan Zhang, Kaidi Xu, Xue Lin, Suman Jana, Cho{-}Jui Hsieh, and
  J.~Zico Kolter.
\newblock Beta-crown: Efficient bound propagation with per-neuron split
  constraints for neural network robustness verification.
\newblock In \emph{Proc. of NeurIPS}, 2021.

\bibitem[Wang et~al.(2022)Wang, Albarghouthi, Prakriya, and Jha]{WangAPJ22}
Zi~Wang, Aws Albarghouthi, Gautam Prakriya, and Somesh Jha.
\newblock Interval universal approximation for neural networks.
\newblock \emph{Proc. of POPL}, 2022.
\newblock \doi{10.1145/3498675}.

\bibitem[Wong \& Kolter(2018)Wong and Kolter]{WongK18}
Eric Wong and J.~Zico Kolter.
\newblock Provable defenses against adversarial examples via the convex outer
  adversarial polytope.
\newblock In \emph{Proc. of ICML}, 2018.

\bibitem[Wong et~al.(2018)Wong, Schmidt, Metzen, and Kolter]{WongSMK18}
Eric Wong, Frank~R. Schmidt, Jan~Hendrik Metzen, and J.~Zico Kolter.
\newblock Scaling provable adversarial defenses.
\newblock In \emph{Proc. of NeurIPS}, 2018.

\bibitem[Xu et~al.(2021)Xu, Zhang, Wang, Wang, Jana, Lin, and
  Hsieh]{XuZ0WJLH21}
Kaidi Xu, Huan Zhang, Shiqi Wang, Yihan Wang, Suman Jana, Xue Lin, and
  Cho{-}Jui Hsieh.
\newblock Fast and complete: Enabling complete neural network verification with
  rapid and massively parallel incomplete verifiers.
\newblock In \emph{Proc. of ICLR}, 2021.

\bibitem[Zhang et~al.(2018)Zhang, Weng, Chen, Hsieh, and Daniel]{ZhangWCHD18}
Huan Zhang, Tsui{-}Wei Weng, Pin{-}Yu Chen, Cho{-}Jui Hsieh, and Luca Daniel.
\newblock Efficient neural network robustness certification with general
  activation functions.
\newblock In \emph{Proc. of NeurIPS}, 2018.

\bibitem[Zhang et~al.(2020)Zhang, Chen, Xiao, Gowal, Stanforth, Li, Boning, and
  Hsieh]{ZhangCXGSLBH20}
Huan Zhang, Hongge Chen, Chaowei Xiao, Sven Gowal, Robert Stanforth, Bo~Li,
  Duane~S. Boning, and Cho{-}Jui Hsieh.
\newblock Towards stable and efficient training of verifiably robust neural
  networks.
\newblock In \emph{Proc. of ICLR}, 2020.

\bibitem[Zhang et~al.(2022)Zhang, Wang, Xu, Li, Li, Jana, Hsieh, and
  Kolter]{ZhangWXLLJHK22}
Huan Zhang, Shiqi Wang, Kaidi Xu, Linyi Li, Bo~Li, Suman Jana, Cho{-}Jui Hsieh,
  and J.~Zico Kolter.
\newblock General cutting planes for bound-propagation-based neural network
  verification.
\newblock In \emph{Proc. of NeurIPS}, 2022.

\end{thebibliography}
\bibliographystyle{iclr2024_conference}

\message{^^JLASTREFERENCESPAGE \thepage^^J}

\ifincludeappendixx
	\clearpage
	\appendix
	
\section{Deferred Proofs on Multivariate Functions} \label{app:theory_multi}
\NetworkFormCoverage*

\begin{Proof}
    Given $\mW_i \RR(\vh_R^{i-1})$, we partition the columns of the weight matrix into $\mW_i = (\mW^+_i | \mW^-_i | \mW^\pm_i)$, depending on whether the associated ReLU is stably active, stably inactive, or unstable, respectively. We thus obtain
    \begin{equation*}
        (\mW^+_i | \mW^-_i | \mW^\pm_i) \RR(\vh_R^{k-1}) = \mW^+_i  \vh_R^{k-1} + \mW^\pm_i \RR(\vh_R^{k-1}).
    \end{equation*}
    We update $\vh_{L,new}^{i-1} = \vh_L^{i-1} + \mW^+_i  \vh_R^{i-1}$, by showing that $\mW^+_i  \vh_R^{i-1}$ is still an $(i-1)$-layer network as follows.
    We recursively update weight matrices $\mW_{k,new} = \mW^+_i \mW_k$ to obtain $\mW^+_i \vh^{k} = \mW^+_i \vh^{k-1} + \mW_{k,new} \RR(\vh_R^{k-1})$ until we have reached $k=1$, where we have $ \mW^+_i \vh^1 =  \mW^+_i\vb_L +  \mW^+_i\mW_{0,L} \vv +  \mW_{1,new} \RR(\vb_R + \mW_{0,R} \vx)$.
\end{Proof}

\begin{restatable}[Simplification of $\RR$ Sums w.r.t. \triangle]{lem}{triangle_sum}\label{thm:triangle_sum}
    Let $\mA \in \R^{n \times 1}$ and $\vw \in \mathbb{R}^n$. Then, we have
    \begin{equation*}
        h(z) = \mA^\top \RR(\vw z) 
        \quad \overset{\Delta}{\rightsquigarrow} \quad 
        h'(z) = \gamma z + \alpha \RR(z),
    \end{equation*}
    where $\gamma = \sum_{\substack{i, w_i < 0}} A_i w_i$ and $\alpha = \sum_{\substack{i, w_i > 0}} A_i w_i - \gamma$.
\end{restatable}

\begin{Proof}
    Both $h$ and $h'$ are CPWL functions with slope change only at $z=0$. Thus they are fully defined by their value at the points $z_i \in \{-1, 0, 1\}$. Hence, we can show that $h$ and $h'$ encode the same function by showing their equivalence on these points: $h(0) = 0 = h'(0)$, $h(-1) = \sum_{\substack{i, w_i < 0}} -A_i w_i = - \gamma = h'(-1)$, and $h(1) = \sum_{\substack{i, w_i > 0}} A_i w_i = \alpha + \gamma$.
    As $\gamma z$ and $\alpha \RR(z)$ are convex/concave, and their \triangle-analysis yields their convex hulls, the pointwise sum of their convex hulls, i.e. the \triangle-analysis of $h'$, recovers the convex hull of $h'$ by \cref{thm:convex_hull_sums} and is thus at least as precise as any convex-relaxation of $h$.
\end{Proof}

\begin{restatable}[Simplification of Composed ReLUs w.r.t. \triangle]{lem}{triangle_comp}\label{thm:triangle_comp}
    We have 
    \begin{equation*}
        h(z) = \RR(\gamma z + \alpha \RR(z)) 
        \quad \overset{\Delta}{\rightsquigarrow} \quad 
        h'(z) = \gamma' z + \alpha' \RR(z),
    \end{equation*}
    where $\gamma' = -\RR(-\gamma)$ and $\alpha' = \RR(\alpha + \gamma) - \gamma'$. 
\end{restatable}

\begin{Proof}
    We observe that $h(z)$ is convex and piecewise-linear for any $z \in [l, u] \subset \R$ with $l < 0 < u$ and a slope change only at $z=0$. Its convex hull is thus spanned by $h(l) = \RR(\gamma l) = h'(l)$, $h(0) = h'(0) = 0$, and $h(u) = \RR((\gamma + \alpha) u) = h'(u)$.
    We further observe that the \triangle-relaxation of $\RR(z)$ and $z$ is their convex hull. 
    Finally, the convex hull of the positive sum of convex functions is equal to the pointwise sum of their individual convex hulls (\cref{thm:convex_hull_sums}). 
    Thus, the triangle-relaxation of $h'(z)$ recovers the convex hull of $h(z)$ and thus the tightest possible convex relaxation.
\end{Proof}

For convex functions $f, g \colon \R \to \R$, we define the convex hull $\bc{H}_f([l,u]) = \{(x,y)\mid x \in [l,u], f(x) \leq y \leq f(l) + \tfrac{f(u) - f(l)}{u-l}(x-l)\}$ over $[l,u] \subset \R$. Further, we define the convex hull sum of $f$ and $g$ on $[l,u]$ to be $\bc{H}_f + \bc{H}_g := \{(x, y' + y'') \mid (x, y') \in \bc{H}_f, (x, y'') \in \bc{H}_g\}$. 

\ConvexHullSums*

\begin{Proof}
    We first show that every point in $\bc{H}_{f+g}$ can be obtained from $\bc{H}_f + \bc{H}_g$. Let $(x, y) \in \bc{H}_{f + g}([l,u])$. Then we have 
    \begin{align*}
        (f + g)(x) \leq &y \leq (f + g)(l) + \tfrac{(f + g)(u) - (f + g)(l)}{u - l}(x - l), \\
        f(x) + g(x) \leq &y \leq f(l) + \tfrac{f(u) - f(l)}{u-l}(x-l) + g(l) + \tfrac{f(u) - f(l)}{u-l}(x-l). 
    \end{align*}
    Then we can find a partition of $y = y' + y''$. We know for sure that there exists $t \in [0,1]$ s.t. 
    \begin{equation*}
        y = (1-t)(f+g)(x) + t((f+g)(l) + \tfrac{(f+g)(u) - (f+g)(l)}{u-l}(x-l)).
    \end{equation*}
    Hence if we pick for example 
    \begin{align*}
        y' &= (1-t)f(x) + t(f(l) + \tfrac{f(u) - f(l)}{u-l}(x-l)) \in \bc{H}_f \\
        y'' &= (1-t)g(x) + t(g(l) + \tfrac{g(u) - g(l)}{u-l}(x-l)) \in \bc{H}_g,
    \end{align*}
    we get immediately that $(x, y) \in \bc{H}_f + \bc{H}_g$.   
    The other direction is immediate. 
\end{Proof}

Using \cref{thm:triangle_comp}, we can show that these networks mapping $\R^2$ to $\R$ can be simplified further:

\NetworkSimplification*
Note that, $h^k$, $h_L^{k-1}$ and $h$ map to $\R$, while $\vh^{k-1}_R$ maps to some $\R^n$. 

\begin{Proof}
    We show a more general result on $\bs{h}^k$ with possibly many output dimensions by induction:

    \emph{Induction Hypothesis:} $\vh^i \overset{\Delta}{\rightsquigarrow} \vb_i + \mW_i \vx + \boldsymbol{\alpha}_i \RR(z)$.

    \emph{Base Case:} $\bs{h}^0(\vx) = \bs{b}_0 + \mW_0 \vx$ satisfies the form $\bs{h}^0(\vx) = \bs{b}_0 + \mW_0 \vx + \boldsymbol{\alpha}_0 \RR(z)$ for $\boldsymbol{\alpha}_0 = \bs{0}$, thus we can replace $\bs{h}^0(\vx)$ by itself.

    \emph{Induction Step:} Using the induction hypothesis, we have $\mW_i \RR(\vh_R^{i-1}) = \mW_i \RR(\vb_{i-1} + \mW_{i-1} \vx + \boldsymbol{\alpha}_{i-1} \RR(z))$, which by \cref{thm:network_form_coverage} only changes its activation state at $z=0$. Since $\RR(0) = 0$, we must have $\vb_{i-1} + \mW_{i-1} \vx = \bs{w}z $ for some $\bs{w}$ (recall that $z$ is the projection of $\bs{x}$ on a hyperplane in the input space). Further, applying \cref{thm:triangle_comp}, we obtain
    \begin{align*}
            \mW_i \RR(\vh_R^{i-1}) 
            =& \mW_i \RR(\bs{w}z + \boldsymbol{\alpha}_{i-1} \RR(z)) \\
            \overset{\Delta}{\rightsquigarrow}& \boldsymbol{\gamma}'_i z + \boldsymbol{\alpha}'_{i} \RR(z) = \vb'_{i} + \mW'_{i} \vx + \boldsymbol{\alpha}'_{i} \RR(z).
    \end{align*}
    Using the induction hypothesis, we can thus rewrite:
    $$\bs{h}^i = \bs{h}^{i-1} + \mW_i \RR(\bs{h}_R^{i-1}) \overset{\Delta}{\rightsquigarrow}  \bs{b} + \mW \vx + \bs{\alpha}_{i} \RR(z).$$
\end{Proof}

\TriangleSingleMax*
\begin{Proof}
    We first constrain our parameters by considering the following:
    \begin{itemize}
        \item For $x = y = 0$, we have $f(0,0) = 0$, leading to $b = 0 = h(0,0)$.
        \item For $x < y$, we have $f(x,y) = y = w_x x + w_y y = h(x,y)$ and thus $w_x = 0$, $w_y = 1$. 
        \item For $x>y$, we have $f(x,y) = x = y + \alpha (x-y) = h(x,y)$ and thus $\alpha = 1$.
    \end{itemize}
    Hence we have $h(x,y) = y + \RR(x-y)$. 
    \begin{align*}
        \begin{rcases}
            0 \\
            x-y
        \end{rcases}
        \leq
        \RR(x-y)
        \leq
        \tfrac{1}{2}(x - y + 1)
    \end{align*}
    Adding $y$ results in the following:
    \begin{align*}
        \begin{rcases}
            y \\
            x
        \end{rcases}
        \leq
        h(x,y)
        \leq
        \tfrac{1}{2}(x + y + 1). 
    \end{align*}
    The maximum of the upper bound is attained at $x = y = 1$, where we get $\tfrac{3}{2}$ which is larger than $\max(x,y) = 1$ for $x, y \in [0,1]$.

\end{Proof}

\TriangleMax*

\begin{Proof}
We will prove this theorem in four steps.
\paragraph*{First -- Locality}
    \begin{enumerate}
        \item There exists a point $(x,y=x)$ with an $\epsilon$-neighborhood $\bc{U}'$ such that one of the following holds for any $\RR(v)$ with input $v = h_v(x,y)$ of the network $h$:
            \begin{itemize}
                \item the ReLU is always active, i.e., $\forall (x,y) \in U, \RR(v) = v$,
                \item the ReLU is never active, i.e., $\forall (x,y) \in U, \RR(v) = 0$, or
                \item the ReLU changes activation state for $x=y$, i.e., $\exists v' \in \R, s.t., \RR(v) = \RR(v' (x-y))$.
            \end{itemize}
        \item   Further, there exists a neighborhood $\bc{U}$ of $(x,y)$ such that the above holds under \triangle-analysis, as it depends continuously on the input bounds and becomes exact for any network when the input bounds describe a point. 
        \item Via rescaling and translation, we can assume that the point $(x,y)$ is at $\mathbf{0}$ and that the neighborhood $\bc{U}$ covers $[-1,1]^2$. 
    \end{enumerate}

\paragraph*{Second -- Network Form} 
On the neighborhood $\bc{U}$, any finite ReLU-network $h$ can, w.r.t. \triangle, be replaced by $\vh^k = \vh^{k-1} + \mW \RR(\vh^{k-1})$ with biases $\vb \in \R^{d_k}$, weight matrices $\mW \in \R^{d_k \times d_{k-1}}$, and $h^0(v) = \vb + \mW \vv$, where all ReLUs change activation state exactly for $x=y$ (\cref{thm:network_form_coverage}).

\paragraph{Third -- Network Simplifications}
We can replace $h^k$ w.r.t. triangle with $b + W_k (x, y)^\top + \alpha_k R(x-y)$ (\cref{thm:network_simplification}). 

\paragraph{Fourth -- Conclusion} Every finite ReLU network can be replaced w.r.t. \triangle with a single layer network of the form $h^1(x,y) = b + \mW (x, y)^\top + \alpha R(x-y)$. However, there exists no such network encoding the $\max$-function such that its \triangle-analysis is precise on the interval $[0,1]^2$ (\cref*{thm:triangle_single_max}).
\end{Proof}

\TriangleImpossibility*

\section{Deferred Proofs on Univariate Functions} \label{app:theory_single}
\subsection{Box}

\StepFunction*

\begin{Proof}
    We prove the theorem by considering the three cases separately:
    \begin{enumerate}
        \item For $x \leq x_0$ we have
        \begin{align*}
            \phi_{x_0, x_1, \beta}(x) 
            &= - \RR(-\tfrac{\beta}{x_1 - x_0}\RR(x-x_0) + \beta) + \beta \\
            &= - \RR(-\tfrac{\beta}{x_1 - x_0} \cdot 0 + \beta) + \beta \\
            &= - \RR(\beta) + \beta \\
            &= - \beta + \beta \\
            &= 0.
        \end{align*}
        \item For $x_0 \leq x \leq x_1$ we have
        \begin{align*}
            \phi_{x_0, x_1, \beta}(x) 
            &= - \RR ( - \tfrac{\beta}{x_1 - x_0} \RR(x-x_0) + \beta) + \beta \\
            &= - \RR ( - \tfrac{\beta}{x_1 - x_0} (x-x_0) + \beta) + \beta \\
            &= \tfrac{\beta}{x_1 - x_0} (x-x_0) - \beta + \beta \\
            &= \tfrac{\beta}{x_1 - x_0} (x-x_0).
        \end{align*}
        \item For $x \geq x_1$ we have
        \begin{align*}
            \phi_{x_0, x_1, \beta}(x)
            &= - \RR(-\tfrac{\beta}{x_1 - x_0}\RR(x-x_0) + \beta) + \beta \\
            &= - \RR(-\tfrac{\beta}{x_1 - x_0} (x-x_0) + \beta) + \beta \\
            &= - 0 + \beta \\
            &= \beta.
        \end{align*}
    \end{enumerate}
\end{Proof}

\BoxPreciseStepFunction*

\begin{Proof}
    Consider the box $[l, u] \subseteq \R$. 
    \begin{align*}
        \phi_{x_0, x_1, \beta}([l,u]) 
        &= - \RR(-\tfrac{\beta}{x_1 - x_0}\RR([l,u]-x_0) + \beta) + \beta \\
        &= - \RR ( - \tfrac{\beta}{x_1 - x_0}  \RR([l - x_0, u- x_0]) + \beta) + \beta \\
        &= - \RR ( - \tfrac{\beta}{x_1 - x_0}  [\RR(l - x_0), \RR(u - x_0)] + \beta) + \beta \\
        &= - \RR ([\tfrac{\beta}{x_1 - x_0} \RR(u - x_0), \tfrac{\beta}{x_1 - x_0} \RR(l - x_0)] + \beta) + \beta \\
        &= - \RR ([\tfrac{\beta}{x_1 - x_0} \RR(u - x_0) + \beta, \tfrac{\beta}{x_1 - x_0} \RR(l - x_0) + \beta]) + \beta \\
        &= - \RR ([\tfrac{\beta}{x_1 - x_0} \RR(u - x_0) + \beta, \tfrac{\beta}{x_1 - x_0} \RR(l - x_0) + \beta]) + \beta \\
        &= - [\RR(\tfrac{\beta}{x_1 - x_0} \RR(u - x_0) + \beta), \RR(\tfrac{\beta}{x_1 - x_0} \RR(l - x_0) + \beta)] + \beta \\
        &= [\RR(\tfrac{\beta}{x_1 - x_0} \RR(l - x_0) + \beta), \RR(\tfrac{\beta}{x_1 - x_0} \RR(u - x_0) + \beta)] + \beta \\
        &= [\RR(\tfrac{\beta}{x_1 - x_0} \RR(l - x_0) + \beta) + \beta, \RR(\tfrac{\beta}{x_1 - x_0} \RR(u - x_0) + \beta) + \beta] \\
        &= [\phi_{x_0, x_1, \beta}(l), \phi_{x_0, x_1, \beta}(u)].
    \end{align*} 
\end{Proof}

\BoxPreciseMonotone*

\begin{Proof}
    W.l.o.g. assume $f$ is monotonously increasing. Otherwise, consider $-f$.
    Let $x_i$ for $i \in \{0, \dots, n\}$ be the set of boundary points of the linear regions of $f$ with $x_0 < \dots < x_n$. 
    We claim that 
    \begin{equation*}
        h(x) = f(x_0) + \sum_{i=0}^{n-1} \phi_{x_i, x_{i+1}, f(x_{i+1}) - f(x_i)}(x)
    \end{equation*}
    is equal to $f$ on $\mathbb{I}$ and that the \ibp-analysis of $h$ is precise. We note that $f(x_{i+1}) - f(x_i) > 0$. 

    We first show $f = h$ on $\mathbb{I}$. For cach $x \in \mathbb{I}$ pick $i \in \{1, \dots, n\}$ such that $x_{j-1} \leq x < x_j$. Then 
    \begin{align*}
        h(x) 
        &= f(x_0) + \sum_{i=0}^{n-1} \phi_{x_i, x_{i+1}, f(x_{i+1}) - f(x_i)}(x) \\
        &= f(x_0) + \sum_{i=0}^{j} \phi_{x_i, x_{i+1}, f(x_{i+1}) - f(x_i)}(x) \\
        &= f(x_0) + \sum_{i=0}^{j-1} \phi_{x_i, x_{i+1}, f(x_{i+1}) - f(x_i)}(x) + \phi_{x_j, x_{j+1}, f(x_{j+1}) - f(x_j)}(x) \\
        &= f(x_0) + \sum_{i=0}^{j-1} \left[f(x_{i+1}) - f(x_i)\right] + \tfrac{f(x_{j+1}) - f(x_j)}{x_{j+1} - x_j}x \\
        &= f(x_j) + \tfrac{f(x_{j+1}) - f(x_j)}{x_{j+1} - x_j}(x - x_j) \\
        &= f(x),
    \end{align*}
    where we used the piecewise linearity of $f$ in the last step.
    
    Now we show that the analysis of \boxd of $h$ is precise. Consider the box $[l, u] \subseteq \mathbb{I}$. We have
    \begin{align*}
        h([l,u]) 
        &= f(x_0) + \sum_{i=1}^n \phi_{x_i, x_{i+1}, f(x_{i+1}) - f(x_i)}([l,u]) \\
        &= f(x_0) + \sum_{i=1}^{n-1} [\phi_{x_i, x_{i+1}, f(x_{i+1}) - f(x_i)}(l), \phi_{x_i, x_{i+1}, f(x_{i+1}) - f(x_i)}(u)] \\
	    &=  [f(x_0) +\sum_{i=1}^{n-1} \phi_{x_i, x_{i+1}, f(x_{i+1}) - f(x_i)}(l), f(x_0) +\sum_{i=1}^{n-1} \phi_{x_i, x_{i+1}, f(x_{i+1}) - f(x_i)}(u)] \\
        &= [h(l), h(u)] \\
        &= [f(l), f(u)]. 
    \end{align*}
\end{Proof}

\subsection{DeepPoly-0}

\ConvModeling*

\begin{Proof}
    First, we show that $\gamma_j = \alpha_j - \alpha_{j-1}$. Evaluating $h(x)$ for $x_j \leq x \leq x_{j+1}$ yields
    \begin{align*}
        h(x) 
        &= 
        b + c x + \sum_{i=1}^{n-1} \gamma_i \RR(\pm_i(x-x_i))
        \\
        &= 
        b + c x + \sum_{i=1}^{n-1} \gamma_i \pm_i (x - x_i) [\pm_i = +, x_i < x] + \sum_{i=1}^{n-1} \gamma_i \pm_i (x - x_i) [\pm_i = -, x_i > x] 
        \\
        &= b + c x + \sum_{i=1}^{n-1} \gamma_i (x - x_i) [\pm_i = +, x_i < x] - \sum_{i=1}^{n-1} \gamma_i (x - x_i) [\pm_i = -, x_i > x] 
        \\
        &= b + c x + \sum_{i=1}^{j} \gamma_i (x - x_i) [\pm_i = +] - \sum_{i=j+1}^{n-1} \gamma_i (x - x_i) [\pm_i = -]. 
    \end{align*}
    The derivative of $h$ evaluated at $x$ for $x_j \leq x \leq x_{j+1}$ is $\alpha_j$:
    \begin{align*}
        \frac{\partial h}{\partial x}(x) 
        &=
        c + \sum_{i=1}^{j} \gamma_i [\pm_i = +] - \sum_{i=j+1}^{n-1} \gamma_i [\pm_i = -] = \alpha_j. 
    \end{align*}
    By choosing $\epsilon$ small enough we can ensure that $x_j + \epsilon \in [x_j, x_{j+1}]$ and $x_j - \epsilon \in [x_{j-1}, x_j]$ and thus
    \begin{align*}
        \alpha_j - \alpha_{j-1} 
        &=  \frac{\partial h}{\partial x}(x_j + \epsilon) - \frac{\partial h}{\partial x}(x_j - \epsilon)  
        \\
        &= \sum_{i=1}^{j} \gamma_i [\pm_i = +] - \sum_{i=j+1}^{n-1} \gamma_i [\pm_i = -] - \sum_{i=1}^{j-1} \gamma_i [\pm_i = +] + \sum_{i=j}^{n-1} \gamma_i [\pm_i = -] \\ 
        &= \gamma_j [\pm_j = +] + \gamma_j [\pm_j = -] \\ 
        &= \gamma_j 
    \end{align*}

    Next, we show that one can pick $\pm_i$ arbitrarily as long as $b$ and $c$ are set appropriately.
    Pick any choice of $\pm_i \in \{-1,1\}$ and set $b$ and $c$ to
    \begin{align*}
        b &:= f(x_0) - x_0 \frac{f(x_1) - f(x_0)}{x_1 - x_0} - \sum_{i=1}^{n-1} \gamma_i x_i [\pm_i = -] = f(x_0) - x_0 \alpha_0 - \sum_{i=1}^{n-1} \gamma_i x_i [\pm_i = -]
        \\
	    c &:= \alpha_0 + \sum_{i=1}^{n-1} \gamma_i [\pm_i = -]. 
    \end{align*}
    We have $h(x) = f(x)$. Indeed: For any $x \in [x_0, x_n]$ pick $j$ s.t. $x \in [x_j, x_{j+1}]$. Then
    \begin{align*}
        h(x) 
        &= 
        b + c x + \sum_{i=1}^{j} \gamma_i (x - x_i) [\pm_i = +] - \sum_{i=j+1}^{n-1} \gamma_i (x - x_i) [\pm_i = -] 
        \\
        &= 
            \underbrace{b  - \sum_{i=1}^j \gamma_i x_i [\pm_i = +] + \sum_{i=j+1}^{n-1} \gamma_i x_i [\pm_i = -]}_\text{offset} 
            + 
            x \underbrace{\left(c + \sum_{i=1}^j \gamma_i [\pm_i = +] - \sum_{i=j+1}^{n-1} \gamma_i [\pm_i = -] \right)}_\text{linear}.
    \end{align*}
    The offset evaluates to 
    \begin{align*}
        &b  - \sum_{i=1}^j \gamma_i x_i [\pm_i = +] + \sum_{i=j+1}^{n-1} \gamma_i x_i [\pm_i = -]
        \\
        &= f(x_0) - x_0 \alpha_0 - \sum_{i=1}^{n-1} \gamma_i x_i [\pm_i = -] - \sum_{i=1}^j \gamma_i x_i [\pm_i = +] + \sum_{i=j+1}^{n-1} \gamma_i x_i [\pm_i = -] 
        \\
		&= f(x_0) - x_0 \alpha_0 - \sum_{i=1}^{j} \gamma_i x_i [\pm_i = -] - \sum_{i=1}^j \gamma_i x_i [\pm_i = +] 
        \\
		&= f(x_0) - x_0 \alpha_0 - \sum_{i=1}^{j} \gamma_i x_i 
        \\
		&= f(x_0) - x_0 \alpha_0 - \gamma_1 x_1 - \gamma_2 x_2 - \dots - \gamma_j x_j 
        \\
		&= f(x_0) - x_0 \alpha_0 - (\alpha_1 - \alpha_0) x_1 - (\alpha_2 - \alpha_1) x_2 - \dots - (\alpha_j - \alpha_{j-1}) x_j 
        \\
		&= f(x_0) - x_0 \alpha_0 + \alpha_0 x_1 - \alpha_1 x_1 + \alpha_1 x_2 - \alpha_2 x_2 - \dots + \alpha_{j-1} x_j - \alpha_j x_j 
        \\
		&= f(x_0) + (x_1 - x_0) \alpha_0 + (x_2 - x_1) \alpha_1 \dots + (x_j - x_{j-1}) \alpha_{j-1} - \alpha_j x_j 
        \\
		&= f(x_0) + (f(x_1) - f(x_0)) + (f(x_2) - f(x_1)) \dots + (f(x_j) - f(x_{j-1})) - \alpha_j x_j 
        \\
		&= f(x_j) - \alpha_j x_j,
    \end{align*}
    where we used that $\gamma_i = \alpha_i - \alpha_{i-1}$ and $\alpha_i = \frac{f(x_{i+1}) - f(x_i)}{x_{i+1} - x_i}$. 
    The linear part evaluates to
    \begin{align*}
        &c + \sum_{i=1}^j \gamma_i [\pm_i = +] - \sum_{i=j+1}^{n-1} \gamma_i [\pm_i = -] 
        \\
		&= \alpha_0 + \sum_{i=1}^{n-1} \gamma_i [\pm_i = -] + \sum_{i=1}^j \gamma_i [\pm_i = +] - \sum_{i=j+1}^{n-1} \gamma_i [\pm_i = -] 
        \\
		&= \alpha_0 + \sum_{i=1}^{j} \gamma_i [\pm_i = -] + \sum_{i=1}^j \gamma_i [\pm_i = +] 
        \\
		&= \alpha_0 + \sum_{i=1}^{j} \gamma_i 
        \\
		&= \alpha_0 + \sum_{i=1}^{j} \alpha_i - \alpha_{i-1} 
        \\
		&= \alpha_i. 
    \end{align*}
    Combining the results, we get 
    \begin{equation*}
        h(x) = f(x_j) - \alpha_j x_j + x \alpha_j = f(x_j) + \alpha_j (x - x_j) = f(x),
    \end{equation*}
    by the piecewise linearity of $f$.
\end{Proof}

\begin{restatable}[\dpO Monotone $\RR$]{lem}{deeppoly_conv_monotone}\label{lem:deeppoly_conv_monotone}
    The \dpO-analysis of the 1-layer $\RR$ network $h(x) = \sum_{i=1}^n \gamma_i \RR(x-x_i)$ yields
    \begin{equation*}
        \begin{rcases}
            h(x),  \\
            h(x_{j-1}) + \alpha_{j-1}(x-x_{j-1})  \\
        \end{rcases}
        \leq 
        h(x) 
        \leq 
        \begin{cases}
            h(x), &\text{if all ReLUs are stable}, \\
            \tfrac{h(u)-h(l)}{u-l}(x-l) + h(l) &\text{otherwise}.
        \end{cases}
    \end{equation*}
    where $x_i \in \R$ s.t. $1 \leq i \leq n$ and $i < p \Rightarrow x_i < x_p$, and $\gamma_i$ are either all $> 0$ or all $< 0$. $j$ is the smallest $i$ such that $x_i \geq l$ and $k$ is the largest $i$ such that $x_i < u$. Thus, \dpO analysis for $h(x)$ is precise.
\end{restatable}

\begin{Proof}
    W.o.l.g. assume $h$ is monotonously increasing. Otherwise, consider $-h$.

    The cases $u < x_1$ and $x_n < l$ are immediate. Choose $j$ as the smallest $i$ such that $x_i \geq l$ and $k$ as the largest $i$ such that $x_i < u$. 
    
    \dpO yields for $\RR(x - x_i)$ on $l \leq x \leq u$ 
    \begin{equation*}
        \begin{rcases}
            x - x_i &\text{if } i \leq j, \\
            0 &\text{if } j < i < k, \\
            0 &\text{if } k \leq i, \\
        \end{rcases}
        \leq
        \RR(x - x_i)
        \leq
        \begin{cases}
            x - x_i &\text{if } i < j, \\
            \tfrac{u - x_i}{u - l}(x - l) &\text{if } j \leq i \leq k, \\
            0 &\text{if } k < i. \\
        \end{cases}
    \end{equation*}
    Thus we have 
    \begin{equation*}
        \sum_{i=1}^{j-1} \gamma_i (x - x_i) \leq h(x) \leq \sum_{i=1}^{j-1} \gamma_i (x - x_i) + \sum_{i=j}^k \gamma_i \tfrac{u - x_i}{u - l}(x - l). 
    \end{equation*}
    The term $\sum_{i=1}^{j-1} \gamma_i (x-x_i)$ can be simplified as follows
    \begin{align*}
        \sum_{i=1}^{j-1} \gamma_i (x - x_i)
        &= \sum_{i=1}^{j-1} \gamma_i x - \sum_{i=1}^{j-1} \gamma_i x_i 
        \\
        &= x \sum_{i=1}^{j-1} (\alpha_i - \alpha_{i-1}) - \sum_{i=1}^{j-1} (\alpha_i - \alpha_{i-1}) x_i 
        \\
        &= x (\alpha_{j-1} - \alpha_{0}) - \sum_{i=1}^{j-1} \alpha_i x_i + \sum_{i=1}^{j-1} \alpha_{i-1} x_i \\
        &= x (\alpha_{j-1} - \alpha_{0}) - \sum_{i=1}^{j-1} \alpha_i x_i + \sum_{i=1}^{j-2} \alpha_{i} x_{i+1} + \alpha_0 x_1 \\
        &= x (\alpha_{j-1} - \alpha_{0}) - \alpha_{j-1} x_{j-1} - \sum_{i=1}^{j-2} \alpha_i x_i + \sum_{i=1}^{j-2} \alpha_{i} x_{i+1} + \alpha_0 x_1 
        \\
        &= x (\alpha_{j-1} - \alpha_{0}) - \alpha_{j-1} x_{j-1} + \sum_{i=1}^{j-2} \alpha_{i} (x_{i+1} - x_{i}) + \alpha_0 x_1 
        \\
        &= \alpha_{j-1} (x-x_{j-1}) + \alpha_0 (x_1 - x) + \sum_{i=1}^{j-2} (h(x_{i+1}) - h(x_{i})) 
        \\
        &= \alpha_{j-1} (x-x_{j-1}) + \alpha_0 (x_1 - x) + h(x_{j-1}) - h(x_1)
        \\
        &= h(x_{j-1}) + \alpha_{j-1} (x-x_{j-1}),
    \end{align*}
    hence we have proven the lower bound. 
    
    Now we consider the upper bound. We evaluate the upper bound at $l$ and $u$. If the two linear upper bounds coincide there, they coincide everywhere:
    \begin{align*}
        x = l \longrightarrow &\sum_{i=1}^{j-1} \gamma_i (l - x_i) + \sum_{i=j}^k \gamma_i \tfrac{u - x_i}{u - l}(l - l)
        \\
        &= \sum_{i=1}^{j-1} \gamma_i (l - x_i) 
        \\
        &= h(l) = \tfrac{h(u)-h(l)}{u-l}(l-l) + h(l),
    \end{align*}
    \begin{align*}
        x = u \longrightarrow &\sum_{i=1}^{j-1} \gamma_i (u - x_i) + \sum_{i=j}^k \gamma_i \tfrac{u - x_i}{u - l}(u - l)
        \\
        &= \sum_{i=1}^{j-1} \gamma_i (u - x_i) + \sum_{i=j}^k \gamma_i (u - x_i)
        \\
        &= \sum_{i=1}^{k} \gamma_i (u - x_i)
        \\
        &= h(u) = \tfrac{h(u)-h(l)}{u-l}(u-l) + h(l),
    \end{align*}
    hence we have proven the upper bound.
\end{Proof}

\DeeppolyConvex*

\begin{Proof}
    The proof works as follows:
    \begin{itemize}
        \item If the function is monotonously increasing, we show that using a local argument $\pm_j = +$ and if the function is monotonously decreasing, we argue $\pm_j = -$. If $f$ has somewhere zero slope, it will be on a piecewise linear region at the boundary of $[x_0, x_n]$, in which case we need $1$ neuron less. 
        \item If the function is non-monotone and has slope zero somewhere, then there are two minima $x^*$ and $x^{**}$ that are also switching points. Hence $f$ is for all $x_j > \tfrac{x^* + x^{**}}{2}$ increasing and for all $x_j < \tfrac{x^* + x^{**}}{2}$ decreasing, so we can reuse the argument from before and need $n-1$ neurons for that. Then we argue separately for the ReLU at $x^*$ and $x^{**}$. 
        \item If the function is non-monotone and has nowhere slope zero, then there exists a unique minimum $x_j$. We then show that there is exactly one splitting of $\gamma_j$ to $\RR(x-x_j)$ and $\RR(-x+x_j)$. 
        \item Finally, we prove that network is precise. 
    \end{itemize}

    We know that there are finitely many switching points $x_i$, $1 \leq i \leq n$. 

    Case 1: $f$ is monotone. W.o.l.g. assume $f$ is monotonously increasing; the proof is similar for monotonously decreasing $f$. Assume that $\pm_j = -$ for some $j$. Then there exists $\epsilon>0$ such that for all $ x \in [x_j - \epsilon, x_j + \epsilon]$, all the ReLUs except $\RR(- x + x_j)$ are stable, i.e., either always 0 or always active. Further, for such inputs, \dpO yields
    \begin{equation*}
        0 \leq \gamma_i \RR(x - x_j) \leq \tfrac{\gamma_j}{2}(- x + x_j + \epsilon).
    \end{equation*}
    As $f$ has no minimum, at least one other $\RR$ need to active at $x_j$: If there would be no other active $\RR$, then $f$ would have a minimum. 
    The active neuron(s) contributes a linear term of the form $\beta x$, $\beta \neq 0$, hence we get for $x_j - \epsilon \leq x \leq x_j + \epsilon$ and some $b \in \R$, \dpO bounds are
    \begin{equation*}
        b + \beta x \leq h(x) \leq b + \beta x + \tfrac{1}{2}(- x + x_j + \epsilon). 
    \end{equation*}
    We know that for $x = x_j$, we get $h(x_j)$ and thus $h(x_j) = b + \beta x_j$.
    The slope of $h$ at $x_j - \epsilon$ is $\beta - \gamma_j$. Hence, we have $h(x_j - \epsilon) = h(x_j) - (\beta - \gamma_j) \epsilon > b + \beta (x_j - \epsilon) = h(x_j) - \beta \epsilon$ since $\gamma_j \epsilon > 0$. This contradicts the assumption that we are precise since the lower bound from \dpO analysis $h(x_j) - \beta$ is unequal to the actual lower bound $h(x_j - \epsilon)$. Hence, we have to have $\pm_j = +$ for all $j$. 

    Case 2: $f$ is not monotone and has slope $0$ somewhere. Then, there are two minima which are also switching points, $x^*$ and $x^{**}$. The argument in Case 1 is a local one, hence we can use the same argument for all $i$ such that $x_i \notin \{x^*, x^{**} \}$, so we only need to consider the cases for $x_j$ and $x_{j+1}$ such that $x_j = x^*$ and $x_{j+1} = x^{**}$. Since $f$ is convex, the only possible case is that $f$ is monotonously decreasing for $x < x_j$ and monotonously increasing for $x > x_{j+1}$, thus we have $\pm_{k}=-$ for $k<j$ and $\pm_{k}=+$ for $k>j+1$. Now we claim that $\pm_j = -$ and $\pm_{j+1} = +$: We have either $(\pm_j, \pm_{j+1}) = (-,+)$ or $(\pm_j, \pm_{j+1}) = (+,-)$. If not, we would not get a unique minimum. The second case also leads directly to a contradiction: Not only would the pre-factors of the two $\RR$ need to coincide, i.e., $\gamma_j = \gamma_{j+1}$ (otherwise one would not have the minimum between them), the analysis of $h$ on $x_j - \epsilon \leq x \leq x_j + \epsilon$ yields (as only the neurons at $x_j$ and $x_{j+1}$ are active there)
    \begin{align*}
        0 &\leq R(x - x_j) \leq \tfrac{1}{2}(x - x_j + \epsilon), \\
        -x+x_{j+1} &= R(-x + x_{j+1}) \\
        b+\gamma_{j+1}(- x + x_{j+1}) &\leq h(x) \leq b+\gamma_{j+1}(-x + x_{j+1}) + \tfrac{\gamma_{j+1}}{2}(x-x_j+\epsilon)
    \end{align*}
     Here the lower bound is $b - \gamma_{j+1} \epsilon < b$, thus imprecise. 

    Case 3: $f$ is not monotone and has slope $0$ nowhere. Then, there is one minimum $x^* = x_j$. For all $x_i \neq x_j$ we can argue as before. So we just need to argue about $x_j$. 
    Assume we only have one $\RR$ involving $x_j$. The only unstable $\RR$ leads to \dpO lower bound $0$, while all others together lead to a linear term $ax+b$ for some $a \ne 0$, thus the overall lower bound from \dpO is $h(x_j) - |a|\epsilon < h(x_j)$. Therefore, such network cannot be precise under \dpO analysis. As one $\RR$ is not enough, we can try two at $x_j$, namely $\gamma_j' \RR(x - x_j)$ and $\gamma_j'' \RR(-x + x_j)$. As around $\gamma_j$ no other $\RR$ is active, it immediately follows that we have $\gamma_j' = \alpha_j$ and $\gamma_j'' = \alpha_{j-1}$.

    Now we finally prove that the network constructed above is precise. Consider the input $l \leq x \leq u$. In the case where $f$ is monotone on $[l,u]$, we have the result immediately by using \cref{lem:deeppoly_conv_monotone} as all $\RR$ with an opposite orientation are inactive. In the case where $f$ is on $[l,u]$ not monotone, the above construction yields
    \begin{equation*}
        h(x) = b + \sum_{i, \pm_i = -} \gamma_i \RR(-(x - x_i)) + \sum_{i, \pm_i = +} \gamma_i \RR(x - x_i). 
    \end{equation*}
    We can apply \cref{lem:deeppoly_conv_monotone} to $\sum_{i, \pm_i = -} \gamma_i \RR(-(x - x_i))$ and $\sum_{i, \pm_i = +} \gamma_i \RR(x - x_i)$ individually to get 
    \begin{align*}
        0 &\leq \sum_{i, \pm_i = -} \gamma_i \RR(-(x - x_i)) \leq \tfrac{h(l) - b}{u - l}(-x + u),
        \\
        0 &\leq \sum_{i, \pm_i = +} \gamma_i \RR(x - x_i) \leq \tfrac{h(u) - b}{u - l}(x - l).
    \end{align*}
    Hence the combined bounds are 
    \begin{equation*}
        b \leq h(x) \leq h(l) + \tfrac{h(u) - h(l)}{u-l} (x-l). 
    \end{equation*}
    Evaluating the upper bounds at $x = l$ and $x = u$ yields $h(l)$ and $h(u)$ respectively, hence the bounds are precise. 
\end{Proof}

\subsection{DeepPoly-1}

\DPoReLU*

\begin{Proof}
    We first prove that $x + \RR(-x) = \RR(x)$. 
    \begin{equation*}
        x + \RR(-x) = \RR(x) - \RR(-x) + \RR(-x) = \RR(x).
    \end{equation*}
    Next we show that the \dpO-analysis of $R(x)$ coincides with the \dpo-analysis of $x + \RR(-x)$. 
    \begin{itemize}
        \item Case $l \leq 0 \leq u$, $x \in [l,u]$: We have for \dpO and $\RR(x)$:
        \begin{equation*}
            0 \leq \RR(x) \leq \tfrac{u}{u-l}(x - l).
        \end{equation*}
        For \dpo and $x + \RR(-x)$ we have:
        \begin{equation*}
            -x \leq \RR(-x) \leq \tfrac{l}{u-l}(x-u),
        \end{equation*}
        Hence
        \begin{equation*}
            0 \leq x + \RR(-x) \leq \tfrac{l}{u-l}(x-u) + x = \tfrac{u}{u-l}(x-l).
        \end{equation*}
        \item Case $0 \leq l \leq u$, $x \in [l,u]$: We have for \dpO and $\RR(x)$:
        \begin{equation*}
            \RR(x) = x. 
        \end{equation*}
        For \dpo and $x + \RR(-x)$ we have:
        \begin{equation*}
            \RR(-x) = 0,
        \end{equation*}
        Hence 
        \begin{equation*}
            x + \RR(-x) = x.
        \end{equation*}
        \item Case $l \leq u \leq 0$, $x \in [l,u]$: We have for \dpO and $\RR(x)$:
        \begin{equation*}
            \RR(x) = 0. 
        \end{equation*}
        For \dpo and $x + \RR(-x)$ we have:
        \begin{equation*}
            \RR(-x) = -x,
        \end{equation*}
        Hence 
        \begin{equation*}
            x + -x = 0.
        \end{equation*}
    \end{itemize}

    To show the opposite, we only need to prove for the case $l \le 0 \le u$ since \dpO and \dpo are identical when there is no unstable $\RR$. 
    For \dpo and $x + \RR(-x)$ we have:
    \begin{equation*}
        0 \leq \RR(-x) \leq \tfrac{l}{u-l}(x-u),
    \end{equation*}
    Hence
    \begin{equation*}
        x \leq x + \RR(-x) \leq \tfrac{l}{u-l}(x-u) + x = \tfrac{u}{u-l}(x-l).
    \end{equation*}
\end{Proof}

\DPoClass*

\begin{Proof}
    The follows immediately with \cref{thm:dp1_ReLU} and the technique presented in \cref{thm:deeppoly_convex}.
\end{Proof}

\subsection{Triangle}

\TriangleConv*

\begin{Proof}
    The \triangle analysis of a $\RR$ over some input range $l \leq x \leq u$ results in the convex hull of $\RR$ on that range. With that, we can apply the \cref{thm:convex_hull_sums} over the network $h$ from \cref{lem:conv_modeling} modeling $f$: 
    \begin{equation*}
        h(x) = b + cx + \sum_{i=1}^{n-1} \gamma_i \RR(\pm_i (x - x_i))
    \end{equation*}
    This is regardless of the choice of $\pm_i$ a sum of convex functions ($\gamma_i > 0$). With \cref{thm:convex_hull_sums} we get that the \triangle analysis of $h$ results in the convex hull of $h$, and thus is precise. 
\end{Proof}

\subsection{Multi-Neuron-Relaxations}

\begin{restatable}[Multi-Neuron Precision]{thm}{multi_neuron}\label{thm:multi_neuron}
    For every $f \in \cpwl(\I, \R)$, there exists a single layer ReLU network $h$ encoding $f$, such that its multi-neuron analysis (considering all ReLUs jointly) is precise.
\end{restatable}

\begin{Proof}
    As the multi-neuron relaxation yields the exact convex hull of all considered neurons in their input-output space, it remains to show that every $f \in \cpwl(\I, \R)$ can be represented using a single-layer ReLU network.

    Recall that every $f \in \cpwl(\I, \R)$ can be defined by the points $\{(x_i, y_i)\}_i$ s.t. $x_i > x_{i-1}$ (\cref{def:cpwl_function}). 
    We set $h_1(x) = (x-x_0) \tfrac{y_1-y_0}{x_1 - x_0} + y_0$ and now update it as follows:
    $$h_{i+1}(x) = h_{i}(x) + \left(\frac{y_{i+1}-y_{i}}{x_{i+1}-x_{i}} - \frac{y_{i}-y_{i-1}}{x_i-x_{i-1}} \right) \RR(x-x_i)$$
    We observe that $\RR(x-x_i) = 0$ for all $x_j$ such that $j \leq i$. As $h(x)$ is CPWL, it is now sufficient to show that $h(x_i) = y_i, \forall i$. 
    
    \begin{align*}
        h(x_j) =& \; (x-x_0) \tfrac{y_1-y_0}{x_1 - x_0} + y_0 + \sum_{i=1} \left(\frac{y_{i+1}-y_{i}}{x_{i+1}-x_{i}} - \frac{y_{i}-y_{i-1}}{x_i-x_{i-1}} \right) \RR(x-x_i) \\ 
               =&\; (x_j-x_0) \tfrac{y_1-y_0}{x_1 - x_0} + y_0 + x_j \sum_{i=1}^{j-1} \left(\frac{y_{i+1}-y_{i}}{x_{i+1}-x_{i}} - \frac{y_{i}-y_{i-1}}{x_i-x_{i-1}}\right) - \sum_{i=1}^{j-1} \left(\frac{y_{i+1}-y_{i}}{x_i-x_{i-1}} - \frac{y_{i}-y_{i-1}}{x_i-x_{i-1}} \right) x_i\\
                =&\; (x_j-x_0) \tfrac{y_1-y_0}{x_1 - x_0} + y_0 + x_{j} \left(\frac{y_{j}-y_{j-1}}{x_{j}-x_{j-1}} - \frac{y_{1}-y_{0}}{x_{1}-x_{0}}\right) - x_{j} \frac{y_{j}-y_{j-1}}{x_{j}-x_{j-1}} \\
                &\;+  \sum_{i=1}^{j-1} \frac{y_{i+1}-y_{i}}{x_{i+1}-x_{i}} (x_{i+1}-x_i) + x_1 \frac{y_{1}-y_{0}}{x_{1}-x_{0}} + \sum_{i=1}^{j-1} \frac{y_{i}-y_{i-1}}{x_i-x_{i-1}} (x_i - x_i)\\
                =&\; y_{j}
    \end{align*}

\end{Proof}
\fi

\end{document}